%% file: main.tex
\PassOptionsToPackage{expansion=false}{microtype}
\documentclass[arxiv]{apple}
\usepackage{amsmath}
\usepackage{float}
\usepackage{enumerate}
\usepackage{algorithm}
\usepackage{algpseudocode}
\usepackage{amsfonts}
\usepackage{amsthm}

\usepackage{colortbl}
\usepackage{amssymb}
\usepackage{xspace}
\usepackage{wrapfig}
\usepackage{adjustbox}
\usepackage{tabularx}
\usepackage{booktabs}
\usepackage{mathtools}
\usepackage{tikz}
\usepackage{enumitem}
\usepackage{silence}
\usepackage{dsfont}
\usepackage{multirow}
\usepackage{makecell}
\usepackage{xfakebold}
\usepackage{arydshln}
\usepackage{bbm}
\usepackage{titletoc}

\usepackage[colorlinks,
linkcolor=red,
anchorcolor=blue,
citecolor=blue
]{hyperref}
\usepackage{cleveref}

\definecolor{textgray}{HTML}{6E6E73}
\usetikzlibrary{positioning, calc}
\usetikzlibrary{decorations.pathmorphing}

\makeatletter
\patchcmd{\wrong@fontshape}{\@gobbletwo}{}{}{}
\makeatother
\makeatletter
\AtBeginDocument{
  \urlstyle{sf}
  
}
\makeatother

\makeatletter
\newcommand\applefootnote[1]{%
  \begingroup
  \renewcommand\thefootnote{}%
  \renewcommand\@makefntext[1]{\noindent##1}%
  \footnote{#1}%
  \addtocounter{footnote}{-1}%
  \endgroup
}
\makeatother

\numberwithin{equation}{section}
\definecolor{light}{RGB}{125, 125, 125}
\crefname{tcb@cnt@pbox}{code}{code}
\Crefname{tcb@cnt@pbox}{Code}{Code}
\crefname{assumption}{assumption}{assumption}
\Crefname{assumption}{Assumption}{Assumptions}
\crefname{appendix}{appendix}{appendices}
\Crefname{appendix}{Appendix}{Appendices}

\newtcolorbox[auto counter]{pbox}[2][]{
  colback=white,
  title=Code~\thetcbcounter: #2,
  #1,fonttitle=\sffamily,
  fontupper=\sffamily,
  arc=2pt,
  colframe=bgcolor,
  coltitle=fgcolor,
  colbacktitle=bgcolor,
  toptitle=0.25cm,
  bottomtitle=0.125cm
}

\newcommand{\sref}[1]{\S\ref{#1}}
\newcommand{\appref}[1]{Appendix~\ref{#1}}
\newcommand{\pp}{\,pp}

\title{LensVLM: Selective Context Expansion for Compressed Visual Representation of Text}

\author[1*]{Roy Xie}
\author[2]{Dan Friedman}
\author[2]{Donghan Yu}
\author[3*]{Bowen Pan}
\author[4*]{Christopher Fifty}
\author[2]{Jang-Hyun Kim}
\author[2]{Xianzhi Du}
\author[2]{\\Zhe Gan}
\author[2]{Vivek Rathod}
\author[1,2]{Bhuwan Dhingra}

\affiliation[1]{Duke University}
\affiliation[2]{Apple}
\affiliation[3]{OpenAI}
\affiliation[4]{Anthropic}
\affiliation[*]{Work done while at Apple}

\abstract{
Vision Language Models (VLMs) offer the exciting possibility of processing text as rendered images, bypassing the need for tokenizing the text into long token sequences.
Since VLM image encoders map fixed-size images to a fixed number of visual tokens, varying rendering resolution provides a fine-grained compression knob.
However, accuracy deteriorates quickly as compression increases: characters shrink below the vision encoder's effective resolution, making them indistinguishable. To address this, we propose LensVLM, an inference framework and post-training recipe
that enables VLMs to scan compressed images, then selectively expand only the relevant images to their uncompressed form via learned tools.
Building on Qwen3.5-9B-Base, LensVLM maintains accuracy comparable to the full-text upper bound at 4.3$\times$ effective compression and outperforms retrieval-based, text- and visual-compression baselines up to 10.1$\times$ effective compression across seven text QA benchmarks.
LensVLM also generalizes to multimodal document and code understanding tasks, with the accuracy gain over baselines growing as compression increases.
Our analysis validates this approach: training makes visual compression robust to rendering choices, and as compression grows the model increasingly relies on expanded content rather than unreliable visual reading.
The analysis also yields practical tool-choice guidance: text expansion is preferable for rendered text, while high-resolution image expansion suits native documents whose layout cues carry task-relevant information.
}
\metadata[Correspondence]{\sffamily \url{ruoyu.xie@duke.edu}}
\metadata[Code]{\url{https://github.com/apple-aiml-research/ml-lensvlm}}
\metadata[Model]{\url{https://huggingface.co/apple/LensVLM-9B}}

\begin{document}

\maketitle

\vspace{1em}
\section{Introduction}
\label{sec:intro}

\begin{wrapfigure}{r}{0.48\textwidth}
  \vspace{-1.5em}
  \centering
  \includegraphics[width=0.46\textwidth]{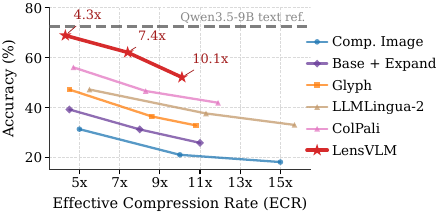}
  \caption{Average QA accuracy vs.\ effective compression rate (ECR) across seven benchmarks. \textsc{LensVLM} maintains accuracy comparable to the full-text upper bound at $4.3\times$ ECR, resulting in the best accuracy--compression trade-off as visual fidelity degrades. 
  }
  \label{fig:teaser}
  \vspace{-1em}
\end{wrapfigure}

A growing line of work treats text not as a sequence of subword tokens but as an image to be consumed by a vision-language model (VLM) \citep{cheng2025glyph, wei2025deepseekocr, xing2025vist, wang2026vtcr1, gao2026zerosense}. 
Under certain rendering configurations, rendered text packs into far fewer visual tokens than its text counterpart. 
This is particularly appealing for long-context applications, since self-attention scales quadratically with sequence length and shorter sequences reduce both compute and memory \citep{vaswani2017attention, tay2022efficient}.

One key advantage of this approach is the flexibility of visual tokenization, as compression options are limited in text space. Retrieval-augmented generation \citep{lewis2020rag} selects relevant passages for the reader but commits a priori to what is retained and requires an offline index, whereas visual tokenization compresses all content while preserving global coverage (detailed comparison in \appref{app:comparison_alternatives}). Token-pruning methods \citep{jiang2023llmlingua, pan2024llmlingua2} can drop tokens, but the compression is discrete and lossy with no way to recover pruned content. More fundamentally, changing text tokenization (e.g., longer subword units) requires vocabulary expansion and costly retraining, and the compression rate is fixed once the tokenizer is chosen. VLMs already process images and text natively, so the compression rate can be tuned continuously by adjusting rendering configuration, without retraining the tokenizer or the model.

 We ask how far this idea can be pushed. Recent work focus on improving VLMs' reading ability \citep{cheng2025glyph}, but it degrades significantly as compression increases. Figure~\ref{fig:teaser} compares different methods using same Qwen3.5-9B-Base backbone except Glyph (GLM-4.1V-9B). The bottleneck is \textit{fidelity}: as compression increases, rendered images shrink, characters become illegible, and model has no way to recover the lost information. Exceeding a certain compression rate, the pixels simply are not readable.

 Therefore, rather than asking the VLM to read perfectly at any compression, we introduce LensVLM\footnote{The name reflects its lens-like ability to focus on compressed visual context and magnify selected regions.}, an inference framework and post-training recipe that augments VLMs with tools to selectively expand parts of their visual context. The model first scans compressed images, then uses a learned tool to expand the relevant regions into uncompressed form (source text or high-resolution images). We argue that compression \textit{should not be} a one-way lossy transform applied uniformly to the context; it \textit{should be} a capability the model internalizes to undo when needed. Unlike retrieval methods that preprocess and index the entire context before any question, LensVLM takes compressed visual tokens as input and expands only the images the model selects on demand. The inference framework works by prompting alone; our post-training recipe teaches smaller models to use it reliably via synthetic traces, SFT, and RL.

 A comprehensive evaluation across seven text QA benchmarks shows that LensVLM maintains accuracy comparable to the full-text upper bound at 4.3$\times$ effective compression rate (ECR) and outperforms baselines up to 10.1$\times$ ECR; ECR measures the total tokens the reader actually processes relative to the source token length (\sref{sec:tool}). Transfer to native documents and code is strongest at high compression (\sref{sec:transfer}).
Our analyses connect these gains to the design choices behind LensVLM. First, training makes visual compression robust to rendering choices, reducing an 18-point accuracy spread across rendering configurations to under 1 point (\sref{sec:rendering_ablation}). Second, controlled tool variants (\sref{sec:tool_variants}) and attention analysis (\sref{sec:attention}) show that readable evidence access drives the gains. Lastly, full end-to-end profiling shows that LensVLM is fastest for long contexts scenarios (\appref{app:efficiency}) and our method is robust to differnt teacher, judge, and rewards (\appref{app:teacher_judge}).

\section{Related Work}
\label{sec:related}

\paragraph{Visual-Text Compression}
A growing line of work compresses text by rendering it as images for VLMs. PIXEL \citep{rust2023pixel} first showed that language modeling over rendered pixels is viable. Glyph~\citep{cheng2025glyph} extends this to long-context reasoning with optimized rendering layouts. VIST~\citep{xing2025vist} introduce dual-path architectures that route distant context through a visual encoder while keeping nearby tokens as text. VTC-R1~\citep{wang2026vtcr1} and AgentOCR~\citep{feng2026agentocr} apply RL to learn what to render. In the text-only space, LLMLingua~\citep{jiang2023llmlingua} and LLMLingua-2~\citep{pan2024llmlingua2} prune tokens directly. Like these methods, we compress text into visual tokens; unlike them, LensVLM introduces an expansion tool to selectively decompress key sections of the input. Existing visual-text compression methods ask the model to answer from a fixed compressed view, so accuracy is bounded by what remains readable at that compression rate. LensVLM instead keeps the compressed view for global scanning and selectively expands relevant regions back to uncompressed form, pushing beyond the limits of visual reading.

\paragraph{Tool-Augmented Reasoning and Document Understanding}
Recent work teaches VLMs to use visual tools such as cropping or zooming ~\citep{su2025twi_survey, zheng2026deepeyes, fan2025grit, wu2026vtoolr1} or select relevant images before reasoning~\citep{cho2024m3docrag, huang2025frag, jain2025simpledoc}. In the document tasks, ARM-Thinker~\citep{ding2025armthinker} combines image retrieval with cropping tools, While others compress document images at the feature level~\citep{hu2024docowl2, thapa2024dragonfly}. Different from previous work, our contribution is a compression formulation rather than a new tool training architecture. Crop/zoom agents such as DeepEyes \cite{zheng2026deepeyes} resolve ambiguity in a \emph{given} image, whereas LensVLM deliberately creates a parameterized lossy encoding of text and makes \textsc{Expand} its functional inverse. LensVLM enables varyious compression rate without tokenizer retraining or corpus re-indexing. Unlike retrieval pipelines that select from an offline index~\citep{xie2025when}, LensVLM preserves a compressed global view and expands regions conditioned on evolving multi-turn context.

\begin{figure}[t]
  \centering
  \includegraphics[width=\linewidth]{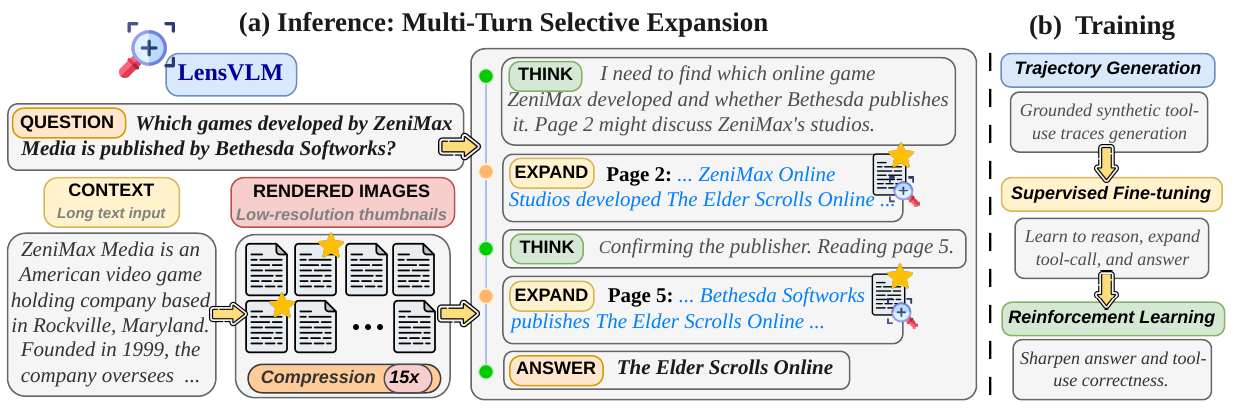}
  \caption{Overview of LensVLM. \textbf{(a)}~Text is rendered into compressed images and fed to the model, which interleaves reasoning with \textsc{Expand} tool to selectively recover relevant images in uncompressed form across multiple turns. \textbf{(b)}~LensVLM is trained through a three-stage pipeline: hard examples are filtered from the base model, synthetic tool-use traces are generated for SFT, and RL further improves answer and tool-use correctness.}
  \label{fig:pipeline}
\end{figure}

\section{Method}
\label{sec:method}
Figure~\ref{fig:pipeline} shows an overview of LensVLM. Text is first rendered into compressed images. The model then selectively expands specific images into uncompressed form via \textsc{Expand} tool. We propose a synthetic data generation pipeline and post‑training recipe to bake this capability into model.

\subsection{Compression: Rendering Text as Images}
\label{sec:rendering}
A compression step uses a deterministic renderer $f$ to convert a text input $\mathcal{D}$ of $N$ tokens into a sequence of $K$ image $\mathcal{V} = f(\mathcal{D}) = (v_1, \ldots, v_K)$. The renderer is parameterized by a rendering configuration (font, geometry, line spacing) that fix text capacity per image. A vision encoder then tokenizes each image $v_k$ into $n_k$ visual tokens, yielding total $\sum_{k=1}^K n_k$ visual tokens fed to the language model. We define the \emph{input compression rate} (ICR) as 
\begin{equation}
C_{\mathrm{in}} \;=\; \frac{N}{\sum_{k=1}^{K} n_k},
\label{eq:compression_rate}
\end{equation}
the ratio of text token to visual token consumed. At $C_{\mathrm{in}}=1$ the model sees one visual token per text token, while $C_{\mathrm{in}} \gg 1$ packs more than one text token into a visual token. 
Higher compression reduces image dimensions and visual tokens per image. Details and examples are in \appref{app:impl_rendering}. 

\subsection{Selective Expansion via Tool Use}
\label{sec:tool}
The model manages its own context through learned $\textsc{Expand}(k)$ tool,  which recover selected compressed images into uncompressed form. Given an image index $k$, $\textsc{Expand}(k)$ returns a representation $\tau_k$ of the corresponding source content: depending on the task, $\tau_k$ may be source text, OCR-extracted text, 
or a high-resolution image (different variants are compared in \sref{sec:tool_variants}).
Let $\pi_\theta$ denote the VLM policy with parameter $\theta$. At inference, the model sees all $K$ images alongside a question $q$ and samples a multi-turn trajectory:
\begin{equation}
y \;\sim\; \pi_\theta(\,\cdot \mid \mathcal{V}, q\,),
\label{eq:rollout}
\end{equation}
where the trajectory $y = (r_1, k_1, \tau_{k_1}, \ldots, r_M, k_M, \tau_{k_M}, a)$ is a sequence of reasoning steps $r_t$, tool calls $\textsc{Expand}(k_t)$ with $k_t \in \{1, \ldots, K\}$, and their responses $\tau_{k_t}$ appended to the context, terminating with a final answer $a$ after $M \leq T$ tool calls. Each turn expands exactly one image; when multiple images are needed, the model expands them sequentially across turns, conditioning each selection on previously expanded content.

\paragraph{Selection-then-Expansion}
Note that the tool introduces no additional information beyond $\mathcal{D}$; it re-presents selected regions in a form that model can read more reliably, recovering information that may be inaccessible after visual compression. Our hypothesis is that compressed images preserve enough corase evidence for the model to identify likely answer-containing regions, even when they do not preserve the token-level content needed for extraction (detailed analysis in \appref{app:theory}). Building on this notion, LensVLM teaches the model to use compressed images for selection, then invoke \textsc{Expand} to access the selected region in uncompressed form. We evaluate this hypothesis empirically: as visual reading degrades under stronger compression, selective expansion remains effective (\sref{sec:ablation_tool}), with trained selection accuracy degrading far more gracefully than visual reading across compression rates (\appref{app:base_model_compression}).

This selection-then-expansion framework is not tied to a particular backbone: it can be applied to frontier VLMs through prompting alone (\appref{app:frontier}), while our post-training recipe teaches a compact 9B model to execute the same behavior end-to-end.

\paragraph{Effective Compression}
Different methods compress the same source text into different numbers of reader-visible tokens, due to differences in vision encoders, retrieval budgets, or tool-call responses. For fair comparson, we use \emph{effective compression rate} (ECR): the ratio of the original text length to the total number of tokens actually processed by the model,
\begin{equation}
C_{\mathrm{eff}} \;=\; \frac{N}{T_{\mathrm{reader}}},
\label{eq:effective_compression}
\end{equation}
where $N$ is the source text token count and $T_{\mathrm{reader}}$ is the total reader input tokens. For LensVLM, $T_{\mathrm{reader}}$ includes both the initial compressed visual tokens and any tokens appended by \textsc{Expand} calls, so $C_{\mathrm{eff}} \leq C_{\mathrm{in}}$. For visual reading without tool, $C_{\mathrm{eff}} = C_{\mathrm{in}}$. We report ECR for all methods, per method details are in \appref{app:ceff_caveats}, and full end-to-end efficiency analysis is in \appref{app:efficiency}.

\subsection{Data Construction}
\label{sec:data}

We construct training data by rendering text from several text QA datasets into images at a fixed ICR and tracking character-level answer spans through pagination to identify the ground-truth evidence images $E^{\star} \subseteq \{1, \ldots, K\}$ for each sample.
Not all samples require tool use: some can be answered directly from the compressed images. Following~\citet{zheng2026deepeyes}, we classify each sample as \textit{easy} (base model answers correctly without tools) or \textit{hard} (tool use needed), determined by running the base VLM on each sample (\appref{app:hard_filter}).

For each hard sample, a synthesis model (Qwen3.5-397B) receives the question $q$, the evidence in its uncompressed form, and the gold answer $a^{\star}$ to generate a synthetic multi-turn trajectory:
\begin{equation}
y^{\star} \;\sim\; \pi_{\text{synth}}(\,\cdot \mid q,\, \mathcal{D}_{E^{\star}},\, a^{\star}\,),
\label{eq:synthetic_trace}
\end{equation}
producing chain-of-thought reasoning interleaved with $|E^{\star}|$ \textsc{Expand} calls. Direct distillation from frontier models is not viable: they produce noisy, hallucination-filled traces when reasoning over highly compressed images, and the \textsc{Expand} behavior is not native to these models. Synthesis sidesteps both issues by providing evidence and answer, reducing the task to generating a coherent reasoning and tool-use trajectory rather than solving visual comprehension under degraded conditions. We also show that a roughly 15$\times$ smaller synthesis model remains effective (\appref{app:teacher_judge}) and replacing annotated evidence spans with synthetic labels only results minimal regressions (\appref{app:evidence_supervision}). The examples can be found in \appref{app:trajectory_examples}.

\subsection{Post-Training}
\label{sec:training}

\paragraph{Supervised Fine-Tuning}

Initializing $\pi_\theta$ from a pretrained VLM and decomposing each synthetic trajectory $y^{\star}$ from \Cref{eq:synthetic_trace} into its $L$ tokens $(y_1^{\star}, \ldots, y_L^{\star})$, the training objective is to minimize the autoregressive loss over $\pi_\theta$ produced tokens:
\begin{equation}
\mathcal{L}_{\mathrm{SFT}}(\theta) = -\sum_{i \in \mathcal{A}} \log \pi_\theta(y_i^{\star} \mid \mathcal{V}, q, y_{<i}^{\star}),
\label{eq:sft_loss}
\end{equation}
where $\mathcal{A} \subseteq \{1, \ldots, L\}$ indexes tokens produced by $\pi_\theta$.
The expanded tool responses $\tau_{k_t}$ remain in the context as inputs but are masked from the loss~\citep{jin2025search}.
We train on the hard subset only since easy samples do not benefit from tool use and including them dilutes the signal. Following~\citet{sun2026modality}, we freeze the vision encoder and visual projector and update only the language model parameters during both SFT and RL (see \appref{app:impl_training} for details).
We denote the resulting parameters $\theta_{\text{SFT}} \in \arg\min_\theta \mathcal{L}_{\mathrm{SFT}}(\theta)$, which initialize the RL stage below.

\paragraph{Reinforcement Learning}

SFT initializes the model with synthetic tool-use trajectories, but these traces reflect the teacher policy distribution. To address this mismatch, we therefore use on-policy RL to let the model learn under its own policy. Starting from the SFT checkpoint, we initialize $\theta \leftarrow \theta_{\text{SFT}}$ and optimize $\pi_\theta$ on a training distribution $\mathcal{P}$ over $(\mathcal{V}, q, a^{\star})$ tuples (images, question, gold answer), maximizing the expected episode reward:
\begin{equation}
\mathcal{J}(\theta) = \mathbb{E}_{(\mathcal{V}, q, a^{\star}) \sim \mathcal{P},\, y \sim \pi_\theta(\cdot \mid \mathcal{V}, q)}\bigl[R(y, a^{\star})\bigr],
\label{eq:rl_objective}
\end{equation}
where $y$ is a model trajectory. Since ground-truth evidence images are not observed during RL, the reward uses only final-answer feedback and an answer-gated tool-use bonus:
\begin{equation}
R(y, a^{\star}) \;=\; {0.7 \cdot c} + {0.3 \cdot c \cdot u},
\label{eq:reward}
\end{equation}
where $c = \mathrm{acc}(y, a^{\star}) \in \{0,1\}$ is answer correctness and $u = \mathds{1}[\textsc{Expand} \in y]$ indicates tool use. We use an LLM judge for $c$ because exact match penalizes valid paraphrases~\citep{wan2025qwenlong,jiang-etal-2025-s3}, although exact-match reward also trains successfully (\appref{app:teacher_judge}). Following \citet{zheng2026deepeyes}, the tool-use bonus is gated by answer correctness. This incentivizes tool use even when the answer could be obtained without it; we accept this bias because the cost of an unnecessary \textsc{Expand} call is small (one extra image) relative to the cost of not reading when needed (a wrong answer). We cap each episode at $T{=}6$ turns; the model averages only 1.3 \textsc{Expand} calls per trajectory (\sref{sec:turns}). We use DAPO~\citep{yu2025dapo} for RL.

\section{Experiments}
\label{sec:experiments}

\subsection{Experimental Setup}
\label{sec:setup}

\paragraph{Datasets \& Models}

We train on NQ~\citep{kwiatkowski2019nq}, HotpotQA~\citep{yang2018hotpotqa}, MuSiQue~\citep{trivedi2022musique}, and HELMET~\citep{yen2024helmet}. In addition to the above benchmarks, we also evaluate on out-of-distribution benchmarks: Qasper~\citep{dasigi2021qasper}, LongBench v1~\citep{bai2024longbench}, and RULER~\citep{hsieh2024ruler}. We evaluate on datapoints with fewer than 32K text tokens; for RULER and LongBench we use the subsets that fall within this budget (\appref{app:data}). We also evaluate on guaranteed-unseen post-cutoff data to check pre-training contamination; the full setup and results are in \appref{app:contamination}.
LensVLM is built on Qwen3.5-9B-Base~\citep{qwen3.5}, a image-text VLM with vision encoder maps each image into a fixed number of visual tokens (72, 48, or 24 at three ICR levels). We train a single LensVLM policy across all compression rates. We study 2B/4B/9B model size scaling in \appref{sec:scaling}. Qwen3.5-397B is the default synthesis model and LLM judge; smaller-teacher and judge-free variants are analyzed in \appref{app:teacher_judge}.

\paragraph{Baselines}

We compare LensVLM against various baselines. As an upper bound, \textbf{Text} feeds the entire document as raw text tokens to the backbone model. \textbf{Comp.\ Image} renders all text as compressed images and feeds them to the backbone model.
\textbf{Base + Expand}, uses the backbone model to read compressed images and predict the evidence image and then expands it.
\textbf{LLMLingua-2}~\citep{pan2024llmlingua2} prunes tokens to match the target compression rate. For image retrieval, we compare with \textbf{ColPali}~\citep{faysse2024colpali}, the state-of-the-art visual retriever over rendered images. Retrieved and the pruned tokens are then fed into the backbone model. All methods above share the same backbone (Qwen3.5-9B-Base) for fair comparison. For model-based methods, we evaluate \textbf{Glyph}~\citep{cheng2025glyph}, a strong same size visual-compression model (GLM-4.1-9B). 
We analyze broader RAG baselines separately in \sref{sec:rag_analysis}. The regimes in which compression and retrieval are preferable are compared in \appref{app:comparison_alternatives}, and more baseline details are in \appref{app:compression_alternatives}.

\paragraph{Evaluation Metrics}
We report QA accuracy as the primary metric, computed by an LLM judge for semantically equivalent answers. We compare different judge's agreement in \appref{app:lm_as_judge}. For each benchmark, we report the unweighted macro-average accuracy across datasets. We also report selection accuracy (Sel. Acc.), the fraction of trajectories where at least one \textsc{Expand} call selects a ground-truth evidence image. We report ECR (\sref{sec:tool}) for fair comparison for all methods. Evaluation uses the same turn budget as training ($T{=}6$).

\subsection{Main Results}
\label{sec:results}

\begin{figure}[t]
\centering
\includegraphics[width=\linewidth]{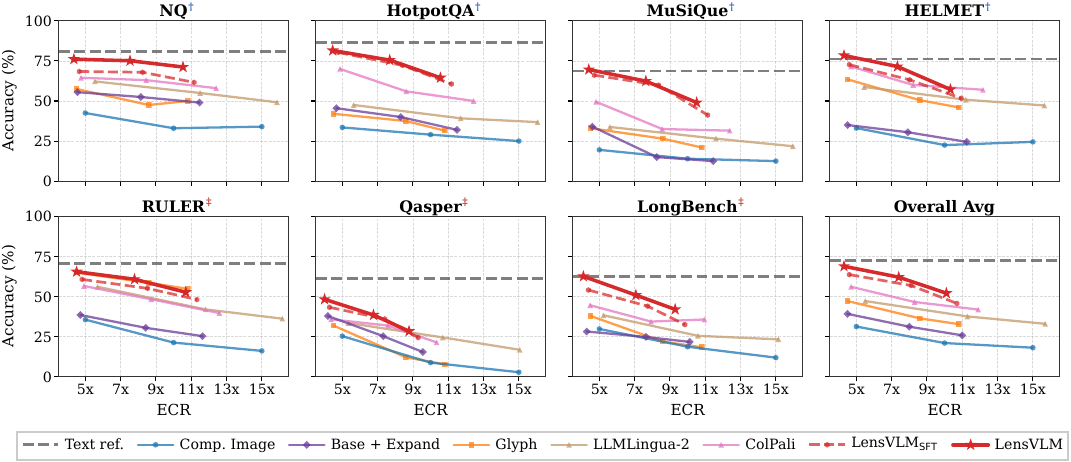}
\caption{LensVLM outperforms or matches all baselines at every compression rate on both in-domain (\textcolor[HTML]{1a60b0}{$\dagger$}) and out-of-domain (\textcolor[HTML]{c0392b}{$\ddagger$}) benchmarks. All methods share the same Qwen3.5-9B-Base reader except Glyph (GLM-4.1V-9B).}
\label{fig:compression_accuracy}

\end{figure}

\paragraph{Text QA Under Compression}

\Cref{fig:compression_accuracy} shows accuracy and ECR trade-offs. Naive visual reading fails: at $5\times$, compressed image drops to 31.3\% against the 72.4\% text upper bound. Glyph partially recovers but is sensitive to different rendering configuration and compression rate~\citep{cheng2025glyph}. Simply predicting and expanding an image with the untrained base model (Base + Expand) yields 39.2\% at 5$\times$, showing that the tool-use augmentation itself provides value even without training. LensVLM maintains 68.9\% at $5\times$ (ECR 4.3$\times$), and achieves the best accuracy--compression trade-off at all three ICRs. Text is the uncompressed accuracy upper bound; under a context budget, LensVLM matches or exceeds all compression baselines at every rate. For retrieval baselines, ECR reflects only the reader's input budget; the indexing pass over the full context is paid offline (\appref{app:comparison_alternatives}). SFT alone lifts accuracy from 31.3\% to 63.7\%, and RL sharpens image selection and reasoning over expanded content, reaching 76.8\%/71.0\%/52.1\% selection accuracy at 5$\times$/10$\times$/15$\times$. We further analyze each training stage in \appref{app:sft_rl}. Full per-dataset QA results are in \Cref{tab:main_results_full} (\appref{app:results}). A contamination-free replication on post-cutoff data confirms comparable gains, averaging +39.7 points across the three ICRs (\appref{app:contamination}).

\label{sec:ablation_tool}
\label{sec:transfer}

\paragraph{Zero-Shot Transfer to Code}
To test whether LensVLM generalizes beyond natural-language text, we evaluate on RepoQA~\citep{liu2024repoqa} and CodeQueries~\citep{sahu2022codequeries}, rendering code with the same pipeline. \Cref{tab:code_results} shows that LensVLM outperforms visual-compression baselines at every compression level without code-specific training. ECR is lower because these contexts are short and the fixed tool-call cost amortizes less. In \appref{app:code_details}, we further show that adding only 8K in-domain code traces substantially closes the remaining gap with negligible regressions on text QA.

\paragraph{Document Understanding with Visual Tools}

We probe transfer to native multimodal document images by training two LensVLM variants with the same procedure and evaluating on MMLongBench-Doc~\citep{ma2024mmlongbench}: one uses an OCR text tool (PaddleOCR-VL~\citep{du2020ppocr}), and one returns a high-resolution image.
We compare both against the base VLM reading original and compressed images (\Cref{tab:doc_results}). At $5\times$, compressed pages remain largely readable, so expansion is often redundant; moreover, zoom selects an evidence page in only 51.5\% of trajectories, and the base VLM may benefit from pretraining exposure to this popular benchmark~\citep{deng2024investigating}. Consequently, compressed-image reading slightly outperforms zoom (51.2\% vs.\ 50.5\%). Tool use helps once fidelity falls: at $15\times$, zoom reaches 41.9\% versus 31.1\% for compressed images. OCR also surpasses the base but trails zoom because it discards layout and visual cues; zoom preserves them at lower ECR. More details are in \appref{app:doc_details}.

\begin{table}[t]
\centering
\begin{minipage}[t]{0.48\textwidth}
\centering
\caption{Code understanding task. LensVLM transfers tool-use to coding tasks without code-specific training.}
\label{tab:code_results}
\vspace{0.5em}
\begin{tabular}{lccc}
\toprule
\textbf{Method} & \textbf{5$\times$} & \textbf{10$\times$} & \textbf{15$\times$} \\
\midrule
Text & 65.6 & 65.6 & 65.6 \\
Comp.\ Image & 26.7 & 6.1 & 3.2 \\
\midrule
Glyph & 31.8 & 9.7 & 7.0 \\
\rowcolor{gray!20} LensVLM & \textbf{38.1} & \textbf{26.1} & \textbf{20.0} \\
\quad \textit{ECR} & 3.2$\times$ & 6.3$\times$ & 8.2$\times$ \\
\quad \textit{Sel. Acc.} & 55.5 & 43.5 & 31.7 \\
\bottomrule
\end{tabular}
\end{minipage}
\hfill
\begin{minipage}[t]{0.48\textwidth}
\centering
\caption{Native document QA. Zoom helps at 10$\times$--15$\times$ but not at 5$\times$, where thumbnails remain readable.}
\label{tab:doc_results}
\vspace{0.5em}
\footnotesize
\begin{tabular}{lccc}
\toprule
\textbf{Method} & \textbf{5$\times$} & \textbf{10$\times$} & \textbf{15$\times$} \\
\midrule
Original Image & 57.5 & 57.5 & 57.5 \\
Comp.\ Image & \textbf{51.2} & 40.3 & 31.1 \\
\midrule
\rowcolor{gray!20}LensVLM + OCR & 45.8 & 38.7 & 32.1 \\
\quad \textit{ECR} & 5.0$\times$ & 9.4$\times$ & 13.5$\times$ \\
\quad \textit{Sel. Acc.} & 51.2 & 39.5 & 33.3 \\
\rowcolor{gray!20}LensVLM + Zoom & 50.5 & \textbf{46.6} & \textbf{41.9} \\
\quad \textit{ECR} & 4.2$\times$ & 7.7$\times$ & 10.3$\times$ \\
\quad \textit{Sel. Acc.} & 51.5 & 42.0 & 36.2 \\
\bottomrule
\end{tabular}
\end{minipage}
\end{table}

\section{Analysis \& Discussion}
\label{sec:analysis}

\subsection{Training Erases Rendering Sensitivity}
\label{sec:rendering_ablation}

\begin{wrapfigure}{r}{0.48\textwidth}
\centering
\includegraphics[width=0.46\textwidth]{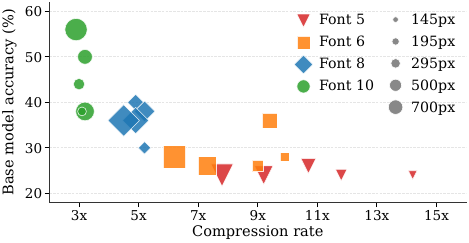}
\captionof{figure}{The base model is highly sensitive to rendering configuration: accuracy spans a 32-point range across 20 configurations.}
\label{fig:rendering_scatter}
\vspace{0.85em}
\captionof{table}{Rendering configuration ablation at matched ${\sim}3\times$ compression. LensVLM eliminates sensitivity.}
\label{tab:rendering_ablation}
\vspace{0.25em}
\small
\begin{tabular}{lcccc}
\toprule
& \textbf{700px} & \textbf{295px} & \textbf{145px} & \textbf{$\Delta$} \\
\midrule
CR & 2.9$\times$ & 3.2$\times$ & 3.1$\times$ & \\
\midrule
Base \ & 56.2 & 50.7 & 38.2 & 18.0 \\
LensVLM \ & 70.3 & 70.4 & 70.8 & 0.5 \\
\bottomrule
\end{tabular}
\end{wrapfigure}

The optimal rendering configuration for text-to-image compression is largely under-explored; finding one often requires exhaustive search over many rendering parameters~\citep{cheng2025glyph,gao2026zerosense}. As shown in \Cref{fig:rendering_scatter}, even at roughly matched compression, the base model remains highly sensitive to rendering: accuracy varies by 32 points across 20 configurations, including an 18-point gap within a single font group. 

To test whether training reduces this sensitivity, we train three variants with the same 10pt font on matched configurations: wide (700px), medium (295px), and tall (145px), all at ${\sim}3\times$ compression (\Cref{tab:rendering_ablation}) using ${\sim}$40K matched data. Before training (Base), accuracy drops monotonically from 56.2\% to 38.2\% as images narrow. After training, all three converge to similar performance, reducing the configuration gap from 18.0 to 0.5 points. This insensitivity likely arises from two sources: trained models specialize to their own rendering configuration, and the text returned by the tool is invariant to how the image was rendered, letting the model rely on expanded content rather than visual legibility (\Cref{sec:attention}). Even at 1$\times$ ECR, no-tool rendering remains 13\pp{} below text, while LensVLM at 5$\times$ ICR is 9.5\pp{} more accurate using 4.3$\times$ fewer reader tokens (\appref{app:high_resolution}), suggesting training removes the need for expensive rendering search, making the configuration a flexible design choice rather than a hyperparameter to optimize.

\subsection{Tool Modality: Text vs.\ Image}
\label{sec:tool_variants}

\begin{wrapfigure}{r}{0.48\textwidth}
\centering
\includegraphics[width=0.46\textwidth]{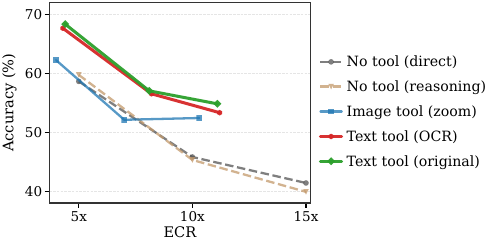}
\caption{Tool modality comparison. Original text outperforms OCR and image zoom; chain-of-thought without tools provides no benefit.}
\label{fig:tool_modality}
\end{wrapfigure}

We trained five model variants with ${\sim}$40K matched samples differing only in tool-call response: (1)~no tool access with a direct answer, (2)~no tool access with reasoning traces, (3)~OCR-extracted text from high-resolution rendered image, (4)~original text, and (5)~a high-resolution version of the selected image (\Cref{fig:tool_modality}). Three findings stand out. \textit{First}, the bottleneck is information access, not reasoning: reasoning without tool access provides no benefit over direct answers, as reasoning over illegible images might introduce noise. \textit{Second}, original text is the best tool response, which is expected. When the original text is unavailable (e.g., multimodal documents), OCR achieves similar performance, making it a practical substitute. \textit{Third}, image zoom consistently underperforms both text tools on rendered text because text tokens provide higher fidelity than visual tokens of the same content. However, on native multimodal document images (\Cref{tab:doc_results}) the ranking reverses: the zoom tool outperforms text tool because text tool discards layout and visual structure that the high-resolution image preserves.

\subsection{Attention Analysis}
\label{sec:attention}

\begin{figure}[t]
  \centering
  \includegraphics[width=\linewidth]{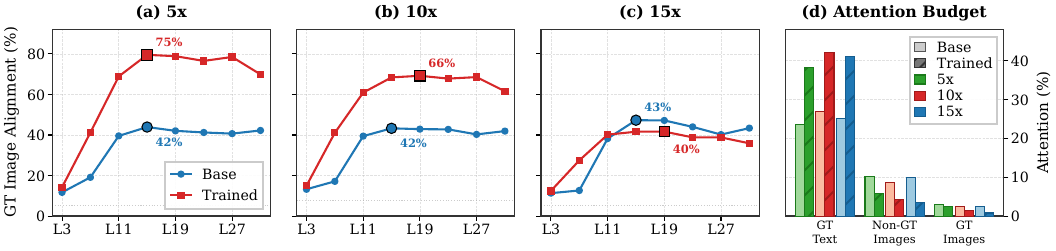}
  \caption{Attention analysis across compression rates. \textbf{(a--c)}~The base model localizes evidence images above chance; post-training sharpens this into predictions. \textbf{(d)}~After training, the model redirects attention from distractor images to expanded text, increasingly so at higher compression.}
  \label{fig:attention}
\end{figure}

To understand what post-training changes, we analyze attention patterns on test samples at three compression rates (Figure~\ref{fig:attention}; setup details in \appref{app:attn_method}).  For each decode step, we compute per-image attention scores by summing head-averaged attention within each image's visual-token range. Panels (a--c) plot the \emph{GT-alignment rate}, the position-debiased fraction of samples whose peak-attention image is a GT evidence image, across the 8 standard-attention layers of Qwen3.5-9B for Base and Trained at each compression preset. Even the base model localizes evidence images well at every compression rate, consistent with the retrieval-head phenomenon in text-only LLMs~\citep{wu2024retrieval}. Training sharpens this diffuse signal into confident predictions at 5$\times$ and 10$\times$ where text remains readable, but cannot improve alignment at 15$\times$ when the visual signal itself is depleted. Panel (d) clarifies how this change is used by showing the \emph{attention budget}, the fraction of total attention assigned to each input segment. After training, the model allocates less attention to images overall, suppresses distractor images more than GT images, and redirects attention to the tool response. This evidence-focused shift becomes stronger as compression increases (full breakdown in \appref{app:attn_shift}), explaining why tool-mediated expansion becomes more valuable as visual fidelity degrades.

  \begin{table}[htbp]
    \centering
    \caption{LensVLM vs.\ retrieval baselines. LensVLM outperforms or matches all retrievers at every compression level using only CPU preprocessing. \textcolor[HTML]{1a60b0}{C}=CPU, \textcolor[HTML]{c0392b}{G}=GPU.}
    \label{tab:rag_baselines}
    \scriptsize
    \begin{tabular}{llccc}
    \toprule
    \textbf{Method} & \textbf{Preproc.} & \multicolumn{3}{c}{\textbf{Accuracy}} \\
    \midrule
    \multicolumn{2}{l}{\textit{Text retrieval}} & \textbf{top-5} & \textbf{top-3} & \textbf{top-2} \\
    BM25 & Idx$^{\textcolor[HTML]{1a60b0}{\mathrm{C}}}$ & 62.5 & 55.7 & 50.8 \\
    BGE-M3 & Idx+Emb$^{\textcolor[HTML]{1a60b0}{\mathrm{C}}+\textcolor[HTML]{c0392b}{\mathrm{G}}}$ & 65.1 & 57.1 & 51.9 \\
    Jina-v4 & Idx+Emb$^{\textcolor[HTML]{1a60b0}{\mathrm{C}}+\textcolor[HTML]{c0392b}{\mathrm{G}}}$ & 64.9 & 57.9 &  \textbf{52.1} \\
    Qwen3-Emb & Idx+Emb$^{\textcolor[HTML]{1a60b0}{\mathrm{C}}+\textcolor[HTML]{c0392b}{\mathrm{G}}}$ & 64.4 & 56.5 & 51.1 \\
    \midrule
    \multicolumn{2}{l}{\textit{Visual retrieval}} & \textbf{top-5} & \textbf{top-3} & \textbf{top-2} \\
    ColPali & Idx+Emb$^{\textcolor[HTML]{1a60b0}{\mathrm{C}}+\textcolor[HTML]{c0392b}{\mathrm{G}}}$ & 56.1 & 46.6 & 41.9 \\
    Jina-v4-MM & Idx+Emb$^{\textcolor[HTML]{1a60b0}{\mathrm{C}}+\textcolor[HTML]{c0392b}{\mathrm{G}}}$ & 63.7 & 54.6 & 50.3 \\
    \midrule
    \multicolumn{2}{l}{\textit{Text-visual compression (ours)}} & \textbf{5$\times$} & \textbf{10$\times$} & \textbf{15$\times$} \\
    LensVLM\textsubscript{SFT} & Rend.$^{\textcolor[HTML]{1a60b0}{\mathrm{C}}}$ & 63.7 & 56.9 & 45.7 \\
    LensVLM & Rend.$^{\textcolor[HTML]{1a60b0}{\mathrm{C}}}$ & \textbf{68.9} & \textbf{62.1} &  \textbf{52.1} \\
    \bottomrule
    \end{tabular}
    \end{table}

\subsection{Empirical Comparison with Retrieval}
\label{sec:rag_analysis}

\Cref{tab:rag_baselines} compares LensVLM with representative text and visual retrievers under the same reader (\appref{app:comparison_alternatives} for the conceptual comparison). To match each compression level, retrieval selects the top-$\lceil K/C_{\mathrm{in}} \rceil$ pages per document (see \appref{app:ceff_caveats} for how compression levels map to retrieval top-$k$). BM25 is already strong (62.5\% at $5\times$); embedding retrievers improve slightly, while visual retrievers (ColPali, Jina-v4-MM) fall in a similar range. LensVLM matches or exceeds every retrieval baseline at every compression level. The \textit{Preproc.} column highlights the structural cost: embedding retrievers require a GPU embedding pass over the full context at index time, whereas LensVLM requires only CPU rendering. We further analyze the selection mechanism through oracle ablations in \appref{app:oracle_selection} and compare with multi-turn agentic baselines in \appref{app:retrieval_diagnostics}.

\section{Conclusion}
\label{sec:conclusion}

We presented LensVLM, an inference framework and post-training recipe for rendered-text compression. Across seven text QA benchmarks, LensVLM maintains performance near the full-text upper bound at $4.3\times$ effective compression, outperforms baselines up to $10.1\times$, and generalizes to documents and code. Comprehensive analysis shows that LensVLM is faster and cheaper than full text on long contexts at high compression.

\bibliographystyle{plainnat}
\bibliography{references}

\newpage
\begin{center}
\section*{Appendix}
\addcontentsline{toc}{section}{Appendix Contents}
\begingroup
\renewcommand{\contentsname}{}
\titlecontents{section}[-1.5em]
  {\normalsize\addvspace{0.5pt}}
  {\thecontentslabel\quad}
  {}
  {\titlerule*[0.25em]{.}\contentspage}[\addvspace{1pt}]
\titlecontents{subsection}[1em]
  {\normalsize\addvspace{0.5pt}}
  {\thecontentslabel\quad}
  {}
  {\titlerule*[0.25em]{.}\contentspage}[\addvspace{0.5pt}]
\startcontents[appendix]
\printcontents[appendix]{}{1}{\setcounter{tocdepth}{2}}
\endgroup
\end{center}

\appendix
\input{appendix}

\applefootnote{\textcolor{textgray}{\sffamily Apple and the Apple logo are trademarks of Apple Inc., registered in the U.S. and other countries and regions.}}

\end{document}

%% file: appendix.tex
\section{Baseline and Compression Method Details}
\label{app:compression_alternatives}

Long-context compression can be approached through text-space methods or visual compression, but each has structural limitations. In text space, token-pruning methods~\citep{jiang2023llmlingua, pan2024llmlingua2} irreversibly discard tokens with no recovery path. Tokenizer-based approaches require an expanded vocabulary and model retraining for each target compression rate, since the rate is baked into the vocabulary. In contrast, visual tokenization requires no vocabulary changes: VLMs already process images natively, so the compression rate is controlled simply by adjusting the rendered image size. Visual compression methods like Glyph~\citep{cheng2025glyph} improve reading from compressed renderings, but still rely on a fixed compressed view with no mechanism to recover information lost to the vision encoder; this helps at lower compression rates but inevitably hits a ceiling at higher compression.

\paragraph{Effective Compression Rate Computation}
\label{app:ceff_caveats}

We report both ICR ($C_{\mathrm{in}}$, \Cref{eq:compression_rate}) and ECR ($C_{\mathrm{eff}}$, \Cref{eq:effective_compression}) in this work. Every baseline family process input images through a different mechanism, producing different achieved ECR. As noted in \appref{app:comparison_alternatives}, retrieval is not compression in the LensVLM sense (the full context is processed at index time); we report ECR based on per-query reader input for fair comparison.

\begin{itemize}[leftmargin=1.5em,itemsep=4pt,topsep=4pt]
  \item \textbf{Text-retrieval} (BM25, BGE-M3, Jina-Text, Qwen3-Emb): Documents are split into chunks matching the text capacity of one rendered image at the corresponding preset. The retriever returns the top $\lceil K / C_{\mathrm{in}} \rceil$ chunks; ECR counts only the retrieved text tokens passed to the reader.

  \item \textbf{Image-retrieval} (ColPali, Jina-Multimodal): The retriever scores rendered images and returns the top $\lceil K / C_{\mathrm{in}} \rceil$; the reader receives the text contained in those images. ECR counts the returned reader text, not the images consumed by the retriever. 

  \item \textbf{Text-compression} (LLMLingua-2): Uses a soft target ICR as a parameter, but because token-level pruning is based on perplexity, the achieved ECR varies by task and can exceed the target.

  \item \textbf{Compressed-image} (Image, Glyph, hybrids): Compression is fixed by rendering preset. For Glyph, the ECR differs because GLM-4.1V produces more visual tokens per image (\Cref{app:model_ecr}); achieved ECR is 4.5$\times$/8.6$\times$/10.8$\times$ at the 5$\times$/10$\times$/15$\times$ presets.

  \item \textbf{Reference baselines}: \emph{Text} feeds the full document ($C_{\mathrm{in}}{=}1$). \emph{Oracle Images} feeds only gold-evidence text ($C_{\mathrm{in}} \approx 15\times$) but presupposes oracle access and is a non-achievable upper bound.

  \item \textbf{LensVLM}: ICR is the rendering preset; ECR is lower because \textsc{Expand} calls append tokens, but adaptive across queries (easy queries retain $C_{\mathrm{eff}} = C_{\mathrm{in}}$). 
\end{itemize}

\paragraph{Model-based visual compression.}
\label{app:model_ecr}
When a model-based baseline uses a different vision encoder, its ECR differs from the nominal ICR. Our presets are designed for Qwen3.5's encoder (\Cref{eq:vt_formula}), which produces $n = \lceil H/16 \rceil \times \lceil W/16 \rceil / 4$ tokens per image. Glyph uses GLM-4.1V-9B-Base, whose encoder has patch size 14 and spatial merge size 2, yielding
\begin{equation}
  n_{\mathrm{GLM}} = \mathrm{round}(H/28) \times \mathrm{round}(W/28)
  \label{eq:glm_vt}
\end{equation}
tokens per image. This produces 25--46\% more visual tokens than Qwen3.5 at our preset dimensions:

\begin{center}
\small
\begin{tabular}{lccccc}
\toprule
\textbf{Preset} & \textbf{Image (W$\times$H)} & \textbf{Qwen3.5 VT} & \textbf{GLM-4.1V VT} & \textbf{Qwen3.5 ECR} & \textbf{Glyph ECR} \\
\midrule
5$\times$ & 256$\times$284 & 72 & 90 & 5.0$\times$ & 4.5$\times$ \\
10$\times$ & 192$\times$252 & 48 & 63 & 10.0$\times$ & 8.6$\times$ \\
15$\times$ & 128$\times$190 & 24 & 35 & 15.0$\times$ & 10.8$\times$ \\
\bottomrule
\end{tabular}
\end{center}

The discrepancy arises because Qwen3.5 uses patch stride 16 with $2{\times}2$ merge (effective 32$\times$32 px/token) while GLM-4.1V uses patch size 14 with $2{\times}2$ merge (effective 28$\times$28 px/token). We report Glyph's achieved ECR in \Cref{tab:main_results_full} (\Cref{app:results}); in general, any model-based baseline should report ECR from its own encoder's tokenization formula.

\subsection{Why Compression over Retrieval${?}$}
\label{app:comparison_alternatives}

Retrieval is the dominant practical alternative for reducing reader input in long-context QA, and \textsc{Expand} can be viewed as a learned retriever in which the VLM itself acts as indexer, selector, and reader in a single end-to-end model. The two paradigms, however, differ along three axes. \emph{Compression mechanism}: retrieval reduces coverage by selecting the top-$\lceil K/C_{\mathrm{in}} \rceil$ of $K$ images at rate $C_{\mathrm{in}}{\times}$ and discarding the rest, whereas LensVLM reduces visual fidelity, rendering all images at lower resolution to preserve global overview. \emph{Preprocessing}: retrieval requires an offline index building on the full context (source text or high-resolution images) with chunking granularity and top-$k$ fixed a priori, while LensVLM needs only CPU rendering and makes \textsc{Expand} selections dynamically at inference, conditioned on the question and prior expansions across turns. \emph{Regime}: retrieval is preferable when evidence is concentrated in a few chunks and indexing cost amortizes across many queries; visual compression is preferable when evidence may be distributed (so top-$k$ risks dropping relevant images), when the source is natively visual with layout cues that text retrievers cannot preserve, or when offline indexing is impractical. The two are complementary: a cheap retriever could prune grossly irrelevant chunks before rendering. Consistent with this distinction, retrieval matches LensVLM on single-hop NQ at 5$\times$, whereas LensVLM leads by 10.6/14.2\pp{} on multi-hop MuSiQue at 5$\times$/10$\times$ (\appref{app:retrieval_diagnostics}).

\subsection{Retrieval Comparsion}
\label{app:retrieval_diagnostics}

\paragraph{Coverage versus fidelity.}
LensVLM uses familiar retrieval and visual-tool components but assigns them a different role: rendering is a deliberate, parameterized lossy encoding, and \textsc{Expand} is its functional inverse. Retrieval instead preserves selected pages exactly but discards the remainder. These failure modes predict different regimes: retrieval is strong when evidence is concentrated, while the compressed global view helps when evidence is distributed across pages.

\begin{table}[H]
\centering
\caption{Retrieval by evidence structure. A strong retriever matches LensVLM on single-hop NQ, whereas LensVLM's advantage grows on multi-hop MuSiQue.}
\label{tab:retrieval_evidence_structure}
\small
\setlength{\tabcolsep}{3.5pt}
\resizebox{\textwidth}{!}{%
\begin{tabular}{llrrrrrr}
\toprule
\textbf{Benchmark} & \textbf{Evidence} & \textbf{LensVLM 5$\times$} & \textbf{Best retr.\ 5$\times$} & \textbf{$\Delta$} & \textbf{LensVLM 10$\times$} & \textbf{Best retr.\ 10$\times$} & \textbf{$\Delta$} \\
\midrule
NQ & 1 hop, concentrated & 76.1 & 76.5 & $-$0.4 & 75.1 & 72.2 & +2.9 \\
MuSiQue & 2--4 hops, distributed & \textbf{69.6} & 59.0 & \textbf{+10.6} & \textbf{62.4} & 48.2 & \textbf{+14.2} \\
\bottomrule
\end{tabular}%
}
\end{table}

At 5$\times$, retrieval slightly exceeds LensVLM on single-hop NQ. On MuSiQue, which requires evidence from two to four pages, LensVLM leads the best retriever by 10.6\pp{} at 5$\times$ and 14.2\pp{} at 10$\times$. Thus the benefit is concentrated where top-$k$ selection is most likely to lose part of the evidence chain. The training-free framework remains separable from this learned policy: prompting alone improves the off-the-shelf 9B backbone by 7.9--10.2\pp{} and Sonnet~4.6 by 4.8--9.0\pp{} (\appref{app:framework_agnostic}).

\paragraph{Agentic baselines.}
We also evaluate the official released checkpoints of Search-R1-7B~\citep{jin2025search}, a text-search agent, and DeepEyes-7B~\citep{zheng2026deepeyes}, a crop/zoom visual agent, on the same seven-benchmark aggregate. Search-R1 indexes source documents at our page granularity and uses its native search budget; DeepEyes receives the same compressed page images as LensVLM and its native zoom tool.

\begin{table}[H]
\centering
\caption{Comparison with released multi-turn agent checkpoints. Search-R1 uses its native text-search budget; DeepEyes receives the same compressed images as LensVLM.}
\label{tab:agentic_retrieval}
\small
\begin{tabular}{lccc}
\toprule
\textbf{Method} & \textbf{5$\times$} & \textbf{10$\times$} & \textbf{15$\times$} \\
\midrule
Search-R1-7B & 46.5 & 46.5 & 46.5 \\
DeepEyes-7B & 26.1 & 20.2 & 15.5 \\
LensVLM & \textbf{68.9} & \textbf{62.1} & \textbf{52.1} \\
\bottomrule
\end{tabular}
\end{table}

Search-R1 is compression-independent because it searches source text at a fixed native budget; it trails LensVLM by 22.4/15.6/5.6\pp{} across the three presets. DeepEyes is the closer visual-tool comparison, but degrades from 26.1 to 15.5 as the rendered text shrinks. LensVLM instead reads at 4.3$\times$/7.4$\times$/10.1$\times$ ECR and remains substantially more accurate, showing that a generic zoom policy does not substitute for learning expansion as the inverse of compression.

\section{End-to-End Efficiency Analysis}
\label{app:efficiency}

We report the complete end-to-end profile of the Qwen3.5-9B reader, including tokenization, rendering, retrieval or pruning, model execution, and tool invocation. We optimize all methods with CUDA graphs and use cross-turn prefix caching and asynchronous rendering for LensVLM. We split evaluation into \emph{short} (7K--32K source tokens) and \emph{long} (32K--140K) contexts as the trade-off depends on context length. We also report \emph{LensVLM (cached)}, where page images are rendered offline. Because \textsc{Expand} operates at page level, rendering is query-independent and can be precomputed per document.

\begin{table}[H]
\centering
\caption{Complete efficiency profile across input compression rates. Latency is seconds/request; KV is peak cache in MiB; throughput is peak requests/s at saturating concurrency; cost is dollars per 1K requests at \$6/GPU-hour. Best is bold.}
\label{tab:efficiency_full}
\scriptsize
\setlength{\tabcolsep}{4pt}
\begin{tabular}{llrrrrrr}
\toprule
\textbf{Metric} & \textbf{Method} & \textbf{S-5$\times$} & \textbf{S-10$\times$} & \textbf{S-15$\times$} & \textbf{L-5$\times$} & \textbf{L-10$\times$} & \textbf{L-15$\times$} \\
\midrule
\multirow{5}{*}{Latency $\downarrow$}
& Full text & \textbf{1.44} & \textbf{1.41} & \textbf{1.39} & \textbf{3.74} & 3.93 & 3.62 \\
& ColPali & 1.59 & 1.55 & 1.76 & 4.58 & 3.72 & 4.63 \\
& LLMLingua-2 & 1.59 & 1.92 & \textbf{1.39} & 4.82 & 4.74 & 4.38 \\
& LensVLM & 1.81 & 1.75 & 1.85 & 4.09 & 2.89 & 2.83 \\
& LensVLM (cached) & 1.77 & 1.72 & 1.81 & 3.80 & \textbf{2.67} & \textbf{2.52} \\
\midrule
\multirow{4}{*}{KV $\downarrow$}
& Full text & 389 & 389 & 398 & 3,044 & 3,047 & 3,116 \\
& ColPali & 91 & 57 & 43 & 622 & 323 & 225 \\
& LLMLingua-2 & \textbf{77} & \textbf{46} & \textbf{30} & 575 & 291 & \textbf{199} \\
& LensVLM & 92 & 67 & 58 & \textbf{515} & \textbf{286} & 221 \\
\midrule
\multirow{5}{*}{Throughput $\uparrow$}
& Full text & \textbf{4.90} & 5.49 & 3.86 & 0.59 & 0.60 & 0.57 \\
& ColPali & 2.43 & 3.19 & 1.90 & 0.40 & 0.51 & 0.39 \\
& LLMLingua-2 & 2.07 & 2.12 & 2.95 & 0.50 & 0.52 & 0.54 \\
& LensVLM & 4.47 & 6.93 & 4.40 & 0.80 & 1.73 & 1.87 \\
& LensVLM (cached) & 4.57 & \textbf{7.05} & \textbf{4.50} & \textbf{0.86} & \textbf{1.87} & \textbf{2.10} \\
\midrule
\multirow{5}{*}{Cost $\downarrow$}
& Full text & \textbf{0.34} & 0.30 & 0.43 & 2.84 & 2.78 & 2.90 \\
& ColPali & 0.69 & 0.52 & 0.88 & 4.21 & 3.25 & 4.25 \\
& LLMLingua-2 & 0.80 & 0.78 & 0.56 & 3.35 & 3.19 & 3.09 \\
& LensVLM & 0.37 & \textbf{0.24} & 0.38 & 2.08 & 0.96 & 0.89 \\
& LensVLM (cached) & 0.36 & \textbf{0.24} & \textbf{0.37} & \textbf{1.94} & \textbf{0.89} & \textbf{0.79} \\
\bottomrule
\end{tabular}
\end{table}

\begin{table}[H]
\centering
\caption{Latency breakdown for representative regimes (seconds/request). CPU includes all host-side preparation; re-prefill is LensVLM's additional forward pass after expansion.}
\label{tab:latency_breakdown}
\scriptsize
\begin{tabular}{llrrrrr}
\toprule
\textbf{Regime} & \textbf{Method} & \textbf{CPU} & \textbf{Prefill} & \textbf{Re-prefill} & \textbf{Decode} & \textbf{Total} \\
\midrule
\multirow{5}{*}{Short, 5$\times$}
& Full text & \textbf{0.04} & 0.19 & -- & 1.21 & \textbf{1.44} \\
& ColPali & 0.45 & \textbf{0.06} & -- & \textbf{1.08} & 1.59 \\
& LLMLingua-2 & 0.29 & \textbf{0.06} & -- & 1.25 & 1.59 \\
& LensVLM & 0.15 & 0.09 & 0.07 & 1.50 & 1.81 \\
& LensVLM (cached) & 0.11 & 0.09 & 0.07 & 1.50 & 1.77 \\
\midrule
\multirow{5}{*}{Long, 15$\times$}
& Full text & \textbf{0.19} & 2.07 & -- & 1.37 & 3.62 \\
& ColPali & 3.17 & 0.11 & -- & \textbf{1.36} & 4.63 \\
& LLMLingua-2 & 2.16 & \textbf{0.10} & -- & 2.13 & 4.38 \\
& LensVLM & 0.88 & 0.21 & 0.13 & 1.61 & 2.83 \\
& LensVLM (cached) & 0.57 & 0.21 & 0.13 & 1.61 & \textbf{2.52} \\
\bottomrule
\end{tabular}
\end{table}

In \Cref{tab:latency_breakdown}, \emph{CPU} includes all host-side preparation before the model runs: tokenization, page rendering for LensVLM, index-and-select for ColPali, and token pruning for LLMLingua-2. \emph{Re-prefill} is the additional forward pass paid by LensVLM's multi-turn inference and does not exist for single-turn methods.6

At short context, full text is the fastest method. However, LensVLM remains useful when a user is willing to trade modest latency for lower token use or memory. At long context, LensVLM is fastest at 10$\times$ and 15$\times$: the main saving is prefill because the reader processes roughly an order of magnitude fewer tokens. At long-context 15$\times$, cached LensVLM adds 0.13\,s of re-prefill but reduces prefill from 2.07 to 0.21\,s; page rendering increases CPU time, while ColPali and LLMLingua-2 pay 3.17 and 2.16\,s for preprocessing.

At long context, LensVLM's KV cache is 5.9--14.1$\times$ smaller than full text. More concurrent requests therefore fit in the same GPU memory, improving throughput and reducing serving cost. At 15$\times$, cached LensVLM increases throughput by 3.7$\times$ and lowers serving cost by 3.7$\times$. These results suggest that LensVLM is best suited to long contexts and multiple queries per document: rendering is query-independent, so the cached variant amortizes it exactly as retrieval amortizes its index. It may not be the right choice for one-off queries over short documents, where full text is faster.

\section{Ablation Analysis}
\label{app:ablations}

\subsection{Higher-Resolution Rendering Without Expansion}
\label{app:high_resolution}

\begin{table}[H]
\centering
\caption{Higher-resolution rendering without \textsc{Expand}. Increasing resolution improves direct visual reading but does not close the modality gap; LensVLM remains more accurate at substantially lower reader cost.}
\label{tab:high_resolution}
\small
\begin{tabular}{lcccc}
\toprule
\textbf{Rendering} & \textbf{W$\times$H (px)} & \textbf{Font} & \textbf{ECR} & \textbf{Acc.} \\
\midrule
Full text (no render) & -- & -- & 1.0$\times$ & \textbf{72.4} \\
Comp.\ Image 1$\times$ & 640$\times$896 & 12\,pt & 1.0$\times$ & 59.4 \\
Comp.\ Image 2$\times$ & 512$\times$704 & 10\,pt & 2.0$\times$ & 48.7 \\
Comp.\ Image 5$\times$ & 256$\times$284 & 8\,pt & 5.0$\times$ & 31.3 \\
Comp.\ Image 10$\times$ & 192$\times$252 & 6\,pt & 10.0$\times$ & 21.0 \\
Comp.\ Image 15$\times$ & 128$\times$190 & 5\,pt & 15.0$\times$ & 18.1 \\
\midrule
LensVLM 5$\times$ & 256$\times$284 & 8\,pt & 4.3$\times$ & \textbf{68.9} \\
LensVLM 10$\times$ & 192$\times$252 & 6\,pt & 7.4$\times$ & 62.1 \\
LensVLM 15$\times$ & 128$\times$190 & 5\,pt & 10.1$\times$ & 52.1 \\
\bottomrule
\end{tabular}
\end{table}

For no-tool rendering, ECR equals ICR because no tool response is appended (\sref{sec:tool}). Raising the resolution to the 1$\times$ limit improves direct visual reading, but remains 13.0\pp{} below text. This gap is therefore not solely a resolution limit: rendered text is underrepresented during VLM pretraining~\citep{cheng2025glyph}. LensVLM at 5$\times$ is 9.5\pp{} more accurate than Comp.\ Image at 1$\times$ while using 4.3$\times$ fewer reader tokens; even LensVLM at 15$\times$ exceeds Comp.\ Image at 2$\times$ (52.1 vs.\ 48.7) with approximately 5$\times$ fewer tokens. Together with the 20-configuration sweep in \sref{sec:rendering_ablation}, these results show that resolution alone provides no setting that is both comparably accurate and cheap.

\subsection{SFT vs.\ RL Training}
\label{app:sft_rl}
\label{sec:ablation_training}

\Cref{tab:training_ablation} isolates the contribution of each training stage. SFT teaches the complete tool-use pipeline: format, selection, and extraction simultaneously, reaching 75\%/65\%/46\% selection accuracy at 5$\times$/10$\times$/15$\times$. RL does not introduce new capabilities but refines the model's judgment about \emph{where} to apply them. Because SFT trains on teacher-generated trajectories, the model inherits the teacher's image-selection distribution rather than learning from its own mistakes. On-policy RL closes this gap: selection accuracy improves most at higher compression where the overview is least legible and selection is hardest (+6.3\pp{} at 10$\times$, +6.2\pp{} at 15$\times$), indicating that RL teaches the model to extract subtler cues from degraded compressed images. RL also improves reasoning over expanded content, adding +5--6\pp{} QA accuracy consistently across all compression rates. 

\begin{table}[H]
\centering
\caption{Training stage ablation. SFT teaches the tool-use format (+32\pp{} at 5$\times$); RL adds +5--6\pp{} QA and further improves selection, especially at higher compression.}
\label{tab:training_ablation}
\vspace{0.5em}

\small
\begin{tabular}{l ccc ccc}
\toprule
& \multicolumn{3}{c}{\textbf{QA Accuracy}} & \multicolumn{3}{c}{\textbf{Selection Accuracy}} \\
\cmidrule(lr){2-4} \cmidrule(lr){5-7}
\textbf{Model} & \textbf{5$\times$} & \textbf{10$\times$} & \textbf{15$\times$} & \textbf{5$\times$} & \textbf{10$\times$} & \textbf{15$\times$} \\
\midrule
Base & 31.3 & 21.0 & 18.1 & 25.8 & 18.0 & 18.3 \\
+ SFT & 63.7 & 56.9 & 45.7 & 75.4 & 64.7 & 45.9 \\
+ SFT + RL (LensVLM) & \textbf{68.9} & \textbf{62.1} & \textbf{52.1} & \textbf{76.8} & \textbf{71.0} & \textbf{52.1} \\
\bottomrule
\end{tabular}
\end{table}

RL does increase the \textsc{Expand} call rate relative to SFT, lowering ECR from 4.6$\times$/8.1$\times$/10.7$\times$ to 4.3$\times$/7.4$\times$/10.1$\times$ at 5$\times$/10$\times$/15$\times$ (\Cref{tab:main_results_full} in \Cref{app:results}). This is a side effect of the tool-use bonus in the reward; however, the additional calls co-occur with improved QA accuracy (+5.2/+5.2/+6.4\pp{}) and selection accuracy (+1.4/+6.3/+6.2\pp{}), indicating that the extra expansions are productive rather than degenerate.

\subsection{Teacher and Reward Dependence}
\label{app:teacher_judge}

The inference framework and post-training recipe are separable: prompting selection then expansion improves the off-the-shelf 9B backbone by 7.9--10.2\pp{} and Sonnet~4.6 by 4.8--9.0\pp{} without training, a teacher, or a judge (\appref{app:frontier}). The dependencies below apply only to the post-training recipe.

\begin{table}[H]
\centering
\caption{Synthesis-model ablation. Replacing the 397B model with a roughly 15$\times$ smaller 27B model preserves most of the gain.}
\vspace{0.5em}
\label{tab:teacher_ablation}
\small
\begin{tabular}{lccc}
\toprule
\textbf{Synthesis model} & \textbf{5$\times$} & \textbf{10$\times$} & \textbf{15$\times$} \\
\midrule
Qwen3.5-397B & \textbf{68.9} & \textbf{62.1} & \textbf{52.1} \\
Qwen3.5-27B & 63.8 & 58.5 & 49.4 \\
$\Delta$ & $-$5.1 & $-$3.6 & $-$2.7 \\
\bottomrule
\end{tabular}
\end{table}

\paragraph{Teacher scale.}
We regenerate traces with Qwen3.5-27B and retrain end to end using the same pipeline. Downsizing the synthesis model costs 2.7--5.1\pp{} (\Cref{tab:teacher_ablation}). This modest gap is expected: synthesis receives the question, uncompressed gold evidence, and answer, and only narrates a selection path rather than solving the compressed QA task.

\paragraph{Reward source.}
The default RL reward uses an LLM judge to recognize semantically correct answers. Replacing it with answer exact match also trains successfully, removing the judge dependency, but yields noisier feedback because valid paraphrases are marked incorrect~\citep{wan2025qwenlong,jiang-etal-2025-s3}. The LLM judge is therefore a reward-quality choice rather than a requirement. Separately, evaluation conclusions are insensitive to judge identity: Qwen3.5-397B and GPT-oss 120B agree on 91.7\% of samples and differ by 0.3\pp{} overall (\appref{app:lm_as_judge}).

\subsection{Dependence on Evidence Supervision}
\label{app:evidence_supervision}

LensVLM separates a training-free inference framework from a recipe for teaching smaller models to use it. The framework requires no evidence supervision (\appref{app:framework_agnostic}); within the recipe, evidence labels seed only SFT trajectory synthesis. RL observes the rendered input, question, and gold answer, but never $E^\star$, and inference observes neither evidence labels nor the gold answer.

Our contamination-free study provides one evidence-annotation-free instantiation: it begins with raw PubMed Central articles and no QA pairs or evidence annotations. Because each question is generated from a unique document chunk, its evidence page is known by construction. The resulting model reaches 86.8 at 5$\times$, against a 92.1 full-text upper bound (\appref{app:contamination}).

We also test the setting where QA pairs exist but evidence spans do not. Qwen3-Embedding-8B and Qwen3-Reranker-8B~\citep{zhang2025qwen3embedding} rank the pages and retain the top six; Qwen3.5-397B~\citep{qwen3.5} then labels evidence given the question and gold answer. We regenerate all trajectories with these synthetic labels and retrain end to end (\Cref{tab:evidence_supervision}).

\begin{table}[H]
\centering
\caption{Evidence-supervision ablation. Synthetic evidence removes the need for annotated evidence spans while preserving most of LensVLM's gain.}
\vspace{0.5em}

\label{tab:evidence_supervision}
\small
\begin{tabular}{lcccc}
\toprule
\textbf{Evidence source} & \textbf{Annotation needed} & \textbf{5$\times$} & \textbf{10$\times$} & \textbf{15$\times$} \\
\midrule
GT evidence & Evidence spans & \textbf{68.9} & \textbf{62.1} & \textbf{52.1} \\
Synthetic evidence & None & 65.2 & 60.3 & 48.8 \\
$\Delta$ & -- & $-$3.7 & $-$1.8 & $-$3.3 \\
\bottomrule
\end{tabular}
\end{table}

Removing annotated evidence costs only 1.8--3.7\pp{}. The annotation-free variant still outperforms every non-retrieval compression baseline at all three rates; at 15$\times$, it reaches 48.8 versus 33.0 for the strongest such baseline. Evidence annotations improve the SFT bootstrap but are not required.

\subsection{Effect of Model Scale}
\label{sec:scaling}

\begin{table}[H]
\centering
\caption{Effect of reader scale at 5$\times$ compression. Tool use is learned at every scale, but 9B yields substantially stronger QA and selection.}
\vspace{0.5em}
\label{tab:scaling}
\small
\begin{tabular}{llcc}
\toprule
\textbf{Size} & \textbf{Stage} & \textbf{QA Acc.} & \textbf{Sel. Acc.} \\
\midrule
\multirow{2}{*}{2B} & Base & 13.2 & -- \\
& LensVLM & 45.5 & 49.4 \\
\midrule
\multirow{2}{*}{4B} & Base & 25.7 & -- \\
& LensVLM & 56.3 & 59.1 \\
\midrule
\multirow{2}{*}{9B} & Base & 31.3 & -- \\
& LensVLM & \textbf{68.9} & \textbf{76.8} \\
\bottomrule
\end{tabular}
\end{table}

The 2B model learns the tool-call syntax but remains weak at both localization and QA. The 4B model reaches 59.1\% selection accuracy, yet QA plateaus at 56.3\%, suggesting that reading and reasoning over expanded content remain bottlenecks. At 9B, selection rises to 76.8\% and QA to 68.9\%, supporting our use of the 9B backbone. Thus the framework is usable below 9B, but reliable end-to-end selection and reasoning benefit substantially from reader capacity.

\subsection{Oracle Image Selection}
\label{app:oracle_selection}

To separate the value of readable text access from the learned selection policy, we evaluate variants on the in-domain test set, where ground-truth evidence images are available (\Cref{tab:oracle_selection}). \emph{Base + Expand} gives the untrained model the same text-access tool as LensVLM; \emph{Oracle Selection} instead returns the gold evidence as uncompressed text, equivalent to perfect top-$k$ retrieval and providing a ceiling.

\begin{table}[H]
\centering
\caption{Access and selection ablation (in-domain, 4 datasets $\times$ 200 samples/preset). Tool access alone helps, but most of LensVLM's gain comes from the learned policy.}
\label{tab:oracle_selection}
\vspace{0.5em}
\small
\begin{tabular}{lccccc}
\toprule
\textbf{Method} & \textbf{Text access} & \textbf{Selection} & \textbf{5$\times$} & \textbf{10$\times$} & \textbf{15$\times$} \\
\midrule
Comp.\ Image (no tool) & No & -- & 32.1 & 24.6 & 24.0 \\
Base + Expand & Yes & untrained & 42.5 & 34.5 & 29.5 \\
LensVLM\textsubscript{SFT} & Yes & SFT & 70.2 & 64.8 & 52.9 \\
LensVLM & Yes & learned & \textbf{76.4} & \textbf{71.1} & \textbf{60.5} \\
\midrule
Img + Oracle Text & Yes & oracle + images & 67.4 & 79.1 & 74.4 \\
\emph{Oracle Selection} & Yes & \emph{oracle} & \emph{80.0} & \emph{80.0} & \emph{80.0} \\
\emph{Text (upper bound)} & -- & -- & \emph{78.1} & \emph{78.1} & \emph{78.1} \\
\bottomrule
\end{tabular}
\end{table}

At 5$\times$, direct visual reading scores 32.1. Adding text access without training contributes 10.4\pp{} (Base + Expand, 42.5), whereas the learned policy contributes a further 33.9\pp{} (LensVLM, 76.4). At 15$\times$, access alone contributes only 5.5\pp{}, while the policy adds 31.0\pp{}. Thus most of the gain comes from learning when and what to expand, not merely exposing the original text.

Oracle Selection is flat at 80.0 because it bypasses selection and returns only gold evidence. It exceeds the full-text upper bound (78.1) because the reader avoids the noisy remainder of the document. LensVLM closes the oracle gap to 3.6\pp{} at 5$\times$ and 8.9\pp{} at 10$\times$ while selecting from low-resolution images. Img + Oracle Text retains all compressed images alongside gold text; its weaker 5$\times$ result indicates distraction from the higher-fidelity visual context.

\subsection{Effect of Multi-Turn Interaction}
\label{app:turns}
\label{sec:turns}

Our method allows the model to call \textsc{Expand} multiple times per query. We ablate the effect of the tool-call budget by running inference once with a generous limit (7 calls) and retroactively evaluating accuracy at each budget $m \in \{1, \ldots, 7\}$: for trajectories that used more than $m$ calls, we force-extract the answer from the response at call $m$. 

\begin{wrapfigure}{r}{0.5\textwidth}
  \centering
  \includegraphics[width=0.48\textwidth]{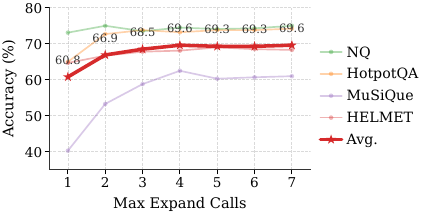}
  \caption{Effect of \textsc{Expand} budget on QA accuracy. One call captures most of the gain (60.8\%); the model self-regulates to ${\sim}$1.3 calls even with generous budgets.}
  \label{tab:turns}
  \end{wrapfigure}

\Cref{tab:turns} shows that a single \textsc{Expand} call provides most of the signal: accuracy reaches 60.8\% after just one call, confirming the effectiveness of enabling tool-use. A second call adds +6.1\pp{} (66.9\%), capturing multi-hop queries that require a second piece of evidence. Returns diminish rapidly after 2 calls: +1.6 (3 calls), +1.1 (4 calls), and $<$0.5 thereafter. In practice, the average trajectory uses only 1.3 \textsc{Expand} calls and nearly all trajectories complete within 5 calls, justifying the default budget of 5 calls ($T{=}6$ turns). MuSiQue benefits most from additional calls, gaining +22\pp{} between 1 and 4 calls (40.3\%$\to$62.5\%) compared to +3.4 for HELMET and +8.2 for HotpotQA, consistent with its multi-hop evidence requirements.

\section{Implementation Details}
\label{app:impl}

\subsection{Rendering}
\label{app:impl_rendering}

\begin{figure}[H]
  \centering
  \begin{subfigure}[t]{0.32\textwidth}
    \vspace{0pt}
    \centering
    \begin{minipage}[t][0.80\linewidth][c]{\linewidth}
      \centering
      \includegraphics[width=\linewidth]{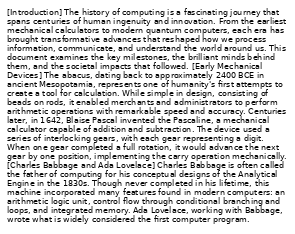}
    \end{minipage}
    \caption{5$\times$ compression}
  \end{subfigure}
  \hfill
  \begin{subfigure}[t]{0.32\textwidth}
    \vspace{0pt}
    \centering
    \begin{minipage}[t][0.80\linewidth][c]{\linewidth}
      \centering
      \includegraphics[width=0.66\linewidth]{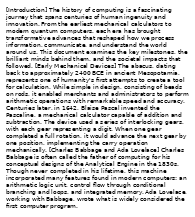}
    \end{minipage}
    \caption{10$\times$ compression}
  \end{subfigure}
  \hfill
  \begin{subfigure}[t]{0.32\textwidth}
    \vspace{0pt}
    \centering
    \begin{minipage}[t][0.80\linewidth][c]{\linewidth}
      \centering
      \includegraphics[width=0.49\linewidth]{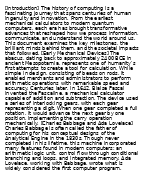}
    \end{minipage}
    \caption{15$\times$ compression}
  \end{subfigure}
  \caption{The same source text rendered at three compression presets. At 5$\times$ (72 VT/img), text is legible to the human eye. At 10$\times$ (48 VT/img) and 15$\times$ (24 VT/img), characters become progressively illegible. The visual token count decreases with compression, where the compression rate reflects both higher text density and fewer visual tokens per image.}
  \label{fig:render_examples}
\end{figure}

We render text into images using PIL with the DejaVuSans font (black text on a white background). Text is word-wrapped to fit the target image width. Each compression preset (\Cref{tab:rendering_presets}) specifies the image width, font size, line spacing, margin, and tokens per image; these parameters fully determine how many characters fit per image, the image height, and thus the visual token count and compression rate. Because tokens per image differs across presets, a given document produces a \emph{different} number of images at each preset (fewer, denser images at higher compression). Example renderings at each compression preset are shown in \Cref{fig:render_examples}.

\paragraph{Vision encoder}
The vision encoder (ViT) maps each image of resolution $H \times W$ to
\begin{equation}
  n = \frac{\lceil H/16 \rceil \times \lceil W/16 \rceil}{4}
  \label{eq:vt_formula}
\end{equation}
visual tokens, where 16 is the grid stride and the division by 4 reflects $2{\times}2$ spatial merging. The encoder operates at native resolution without any additional resizing.

We choose image widths that are multiples of 32 ($= 16 \times 2$) so that $\lceil W/16 \rceil$ and the merged output are exact integers with no rounding, making the visual token count deterministic. Image heights are controlled via the tokens-per-image parameter to also land on grid-aligned values. The resulting per-preset token counts are: 5$\times$: $\lceil 284/16 \rceil \times \lceil 256/16 \rceil / 4 = 18 \times 16 / 4 = 72$; 10$\times$: $\lceil 252/16 \rceil \times \lceil 192/16 \rceil / 4 = 16 \times 12 / 4 = 48$; 15$\times$: $\lceil 190/16 \rceil \times \lceil 128/16 \rceil / 4 = 12 \times 8 / 4 = 24$, yielding a clean 3:2:1 ratio across presets (\Cref{tab:rendering_presets}).

This design means the compression rate is a product of two factors: higher text density per image (more characters packed into a smaller image) \emph{and} fewer visual tokens per image (smaller images produce fewer patches). At 5$\times$, text is legible to the human eye; at 10$\times$, characters are small but distinguishable; at 15$\times$, most text is illegible. We train and evaluate on a mix of presets (5$\times$, 10$\times$, 15$\times$) unless otherwise noted.

For example, for a 10{,}000-token document, the three presets yield: 5$\times$: $\lceil 10{,}000 / 405 \rceil = 25$ images $\times$ 72 VT = 1{,}800 VT (5.6$\times$ compression); 10$\times$: 19 images $\times$ 48 VT = 912 VT (11.0$\times$); 15$\times$: 27 images $\times$ 24 VT = 648 VT (15.4$\times$). Image count stays in a narrow range (19--27) across presets; compression comes primarily from the per-image VT budget shrinking from 72 to 24 (a 3:1 ratio). The model sees a similar number of images at every compression rate, but each becomes smaller and less legible. Tokens per image is approximate and varies with actual content.

\begin{table}[H]
\centering
\caption{Rendering presets. Higher compression packs more text into fewer visual tokens per image, yielding a 3:2:1 VT ratio across presets. Image widths are multiples of 32 for grid-aligned vision-encoder tokenization.}
\vspace{0.5em}
\label{tab:rendering_presets}
\resizebox{\textwidth}{!}{%

\begin{tabular}{lcccccccc}
\toprule
\textbf{Preset} & \textbf{Width (px)} & \textbf{Font (pt)} & \textbf{Line sp.} & \textbf{Margin (px)} & \textbf{Tok/img} & \textbf{Height (px)} & \textbf{VT/img} & \textbf{$\sim$Chars/img} \\
\midrule
5$\times$  & 256 & 8 & 1.15 & 7 & 405 & 284 & 72 & 2{,}000 \\
10$\times$ & 192 & 6 & 1.10 & 6 & 540 & 252 & 48 & 2{,}700 \\
15$\times$ & 128 & 5 & 1.05 & 5 & 378 & 190 & 24 & 1{,}900 \\
\bottomrule
\end{tabular}
}
\end{table}

\subsection{Training}
\label{app:impl_training}

During both SFT and RL, we freeze the vision encoder and visual projector and only update the language model parameters. Recent work~\citep{sun2026modality} shows that adapting the language model is sufficient for text-visual compression reading task, while training the vision encoder yields no additional benefit. This also reduces memory and compute cost, as gradients need not flow through the vision encoder. All training and evaluation are done using 8$\times$ B200 GPUs.

For SFT, we finetune Qwen3.5-9B-Base using LLaMA-Factory~\citep{zheng2024llamafactory} with DeepSpeed ZeRO-3~\citep{aminabadi2022deepspeed}, learning rate $5{\times}10^{-6}$, effective batch size 64 (batch size 2 per GPU with gradient accumulation 4) for 2 epochs on the hard subset. Training data is formatted in ShareGPT multi-turn conversation format with interleaved image tokens.

For RL, we use VERL~\citep{sheng2024verl} to run DAPO~\citep{yu2025dapo} starting from the SFT checkpoint on 15K samples subsampled from the remaining hard samples and easy samples, with learning rate $10^{-7}$, batch size 32, 16 rollouts per prompt, temperature 0.8, up to 6 tool calls per episode, and no entropy regularization. Our scale study spans 2B--9B (\appref{sec:scaling}) due to compute budget. Our generic inference framework and post-training recipe should generalize to larger models, we leave this to future work.

We use Qwen3.5-9B-Base~\citep{qwen3.5} as our backbone VLM. It couples a 9B-parameter language decoder with a vision encoder (ViT with spatial merging) that maps each image into a resolution-dependent number of visual tokens (\Cref{tab:rendering_presets} in \Cref{app:impl_rendering}). We select Qwen3.5-9B for three reasons: (1) it natively supports interleaved image-text inputs, which is required for our multi-turn tool-use format; (2) its vision encoder handles variable-resolution images without resizing, preserving rendered-text fidelity; and (3) the scale study shows substantially stronger QA and selection at 9B than at 2B or 4B (\appref{sec:scaling}).

\subsection{Evaluation}
\label{app:impl_eval}

We serve all models for evaluation using vLLM~\citep{kwon2023vllm}. Models are allowed up to 6 turns with greedy decoding (temperature 0.0) and max response tokens of 2{,}048 per turn. At each turn, the model may invoke the \textsc{Expand} tool; the tool response is appended to the conversation and the model continues until it produces a final answer or exhausts the turn budget. The LLM judge (\sref{sec:setup}) is hosted separately via vLLM and scores each response with greedy decoding.

\begin{table}[H]
  \centering
  \caption{Judge model agreement. Qwen3.5-397B and GPT-oss 120B differ by only 0.3\pp{} overall ($\kappa = 0.81$), confirming that our conclusions are judge-independent.}
  \label{tab:judge_correlation}
  \vspace{0.5em}
  \small
  \begin{tabular}{lccccc}
  \toprule
  \textbf{Compression} & \textbf{Qwen3.5-397B Acc.} & \textbf{GPT-oss 120B Acc.} & \textbf{$\Delta$} & \textbf{Agreement} & \textbf{Cohen's $\kappa$} \\
  \midrule
  5$\times$  & 68.9 & 68.1 & $-$0.8 & 92.7\% & 0.81 \\
  10$\times$ & 62.1 & 62.8 & +0.7 & 91.5\% & 0.80 \\
  15$\times$ & 52.1 & 51.2 & $-$0.9 & 90.8\% & 0.81 \\
  \midrule
  \textbf{Overall} & 61.0 & 60.7 & $-$0.3 & 91.7\% & 0.81 \\
  \bottomrule
  \end{tabular}
  \end{table}
  
\subsection{Judge Model Agreement}
\label{app:lm_as_judge}

We evaluate two judges' agreement on our main text QA results: GPT-oss 120B~\citep{agarwal2025gpt} and Qwen3.5-397B. Both judges use identical prompts (\Cref{app:prompt}) and greedy decoding on the same (question, gold answer, model prediction) triples across all three compression rates. The two judges differ by only 0.3\pp{} overall with inter-judge agreement of 91.7\% (\Cref{tab:judge_correlation}). This establishes insensitivity to judge identity; exact-match reward can remove the training judge entirely (\appref{app:teacher_judge}).

\section{Information-Theoretic Perspective}
\label{app:theory}
We discuss the design motivation behind LensVLM from an information-theoretic perspective.

\paragraph{Information can be lost at rendering and at encoding.}
Rendering maps text $\mathcal{D} = (t_1, \ldots, t_N)$ into images $\mathcal{V} = f(\mathcal{D})$ through a deterministic renderer $f$, and the model observes visual tokens $\mathcal{T} = g(\mathcal{V})$ via the vision encoder $g$. Determinism does not imply injectivity: at sufficiently aggressive compression, two different strings can render to indistinguishable pixel images (e.g., \Cref{fig:render_examples}(c) in \Cref{app:impl_rendering}). Applying the data processing inequality through both stages,
\begin{equation}
  I(a^{\star};\, \mathcal{T} \mid q) \;\le\; I(a^{\star};\, \mathcal{V} \mid q) \;\le\; I(a^{\star};\, \mathcal{D} \mid q),
  \label{eq:info_loss}
\end{equation}
so failures under compression can arise either because the renderer destroys answer-relevant information or because the vision encoder cannot recover it. The \textsc{Expand} tool does not introduce information beyond the source: the returned content $\tau_k$ is a deterministic function of $\mathcal{D}$. Its role is representational; it presents a selected region in a form that is easier for the model to read (source text or a high-resolution image) rather than asking the model to extract all fine-grained content from compressed visual tokens.

\paragraph{Selection and extraction have different targets.}
Image selection and answer extraction ask the model to predict different variables: selection identifies an evidence set $E^{\star} \subseteq \{1, \ldots, K\}$ (a single image for single-hop queries, multiple for multi-hop), while extraction predicts an answer string $a^{\star}$ over a much larger, content-dependent space. This asymmetry suggests (but does not prove) that compressed visual tokens may preserve enough coarse cues for selection even when they do not support reliable character-level reading. Whether this holds is a property of the task distribution and model rather than a theorem: in corner cases such as a factoid whose only clue is a short phrase on one image, selecting the correct image may require reading nearly the same text needed to answer the question.

\paragraph{When selective expansion helps.}
The preceding points suggest when selective expansion should help. Let $\bar{D}(\pi)$ denote the expected task error under policy $\pi$, marginalizing over all subsets $S \subseteq \{1,\ldots,K\}$ that $\pi$ may select for expansion. Conditioning on whether all evidence images are included,
\begin{equation}
  \bar{D}(\pi)
  \;=\;
  p_\pi \, D_{\mathrm{hit}}
  +
  (1-p_\pi) \, D_{\mathrm{miss}},
  \label{eq:distortion_decomp}
\end{equation}
where $p_\pi = P_\pi(E^{\star} \subseteq S)$, $D_{\mathrm{hit}}$ is the expected error when all evidence images are expanded, and $D_{\mathrm{miss}}$ is the error when at least one is not (for multi-hop queries with $|E^\star|>1$, this averages over partial-retrieval cases). Let $D_{\mathrm{no}}(\rho)$ be the error of answering directly from compressed images at compression rate $\rho$. The benefit of selective expansion is
\begin{equation}
  D_{\mathrm{no}}(\rho) - \bar{D}(\pi)
  =
  p_\pi \bigl(D_{\mathrm{no}}(\rho) - D_{\mathrm{hit}}\bigr)
  -
  (1-p_\pi)\bigl(D_{\mathrm{miss}} - D_{\mathrm{no}}(\rho)\bigr).
  \label{eq:benefit}
\end{equation}
This makes the tradeoff explicit: tool use helps when the gain from correctly expanding all evidence images outweighs the penalty from missing evidence or being misled by expanded distractors. The decomposition is exact but does not require $D_{\mathrm{hit}}$ to match the full-text error or $D_{\mathrm{miss}}$ to match the no-tool compressed-image error; expanding the wrong images can hurt if the model over-focuses on irrelevant tool responses.

The perspective predicts a useful compression regime: tool use helps when direct reading degrades faster than selection accuracy. Our main results confirm LensVLM operates in this regime across all studied compression rates (\Cref{fig:compression_accuracy} in \sref{sec:results}), with the advantage growing as compression increases. The pattern is model-dependent: the untrained base model's selection degrades much faster (\Cref{app:base_model_compression}), and attention shifts increasingly toward tool-response tokens at higher compression (\sref{sec:attention}).

\paragraph{Empirical verification.}
\label{app:base_model_compression}
We evaluate whether the untrained base model can use compressed images for evidence localization by comparing QA and selection accuracy across compression rates on the standard multi-preset test set (\Cref{app:data}).

For QA accuracy, the base model receives all compressed images and answers the question directly. For selection accuracy, it receives the same images but is instead prompted to identify which image contains the answer. Both tasks use greedy decoding. QA is scored by the LLM judge, while selection is scored by whether the predicted image matches any ground-truth evidence image. For multi-hop queries (MuSiQue, 2 -- 4 evidence images), this ``any-hit'' criterion is deliberately lenient; QA accuracy remains the primary metric because answering correctly requires retrieving evidence across multiple \textsc{Expand} calls. \Cref{tab:training_ablation} (\Cref{sec:ablation_training}) shows that, without training, the base model's direct QA accuracy falls from 31.3\% at 5$\times$ to 18.1\% at 15$\times$ (42\% relative drop). Selection accuracy is substantially lower: the untrained model reaches only 25.8\% at 5$\times$ and drops to 18.0\% at 10$\times$ (30\% relative degradation, over the same range where QA drops 33\%), indicating that the base model cannot reliably localize evidence from compressed images.

These results show that the information-theoretic argument is not sufficient on its own. Although selection is a plausible lower-dimensional target than extraction, the pretrained model does not already know how to use compressed images for localization. Post-training changes this behavior substantially: after SFT+RL, selection accuracy reaches 76.8\%/71.0\%/52.1\% at 5$\times$/10$\times$/15$\times$ (\Cref{tab:training_ablation} in \Cref{sec:ablation_training}). Thus, selection from compressed images is learnable, but it is not guaranteed by information-theoretic considerations alone.

\section{A Model-Agnostic Inference Framework}
\label{app:framework_agnostic}
\label{app:frontier}

The selection-then-expansion framework (scan compressed images, identify the relevant image, expand to uncompressed form) is not tied to a particular model. To quantify the framework's standalone value, we apply it to two model families without any task-specific training: (1)~an open-weight 9B VLM (Qwen3.5-9B-Base) and (2)~a frontier model (Sonnet~4.6~\citep{anthropic2026sonnet46}). In both cases, the model receives compressed images, predicts which image contains the answer, and then receives that image's content as uncompressed text before generating its final answer.

\begin{figure}[H]
  \centering
  \includegraphics[width=0.65\textwidth]{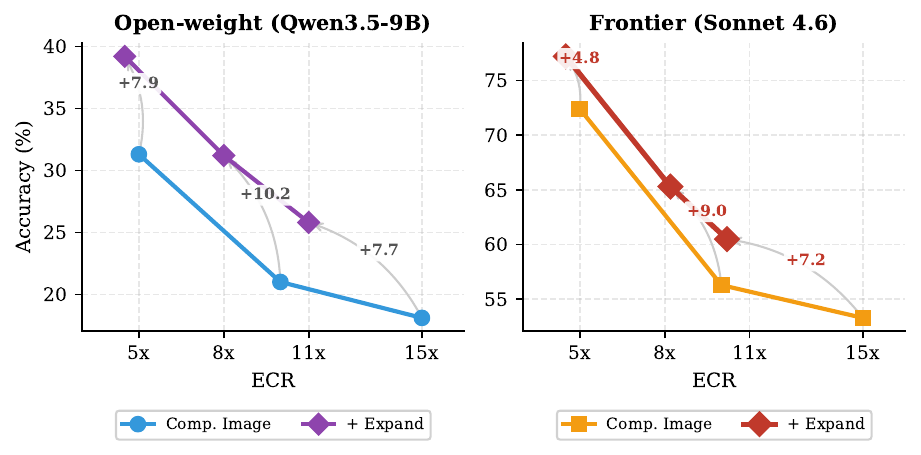}
  \caption{The \textsc{Expand} tool improves accuracy for both open-weight (9B) and frontier (Sonnet~4.6) models at every compression level without any fine-tuning.}
  \label{fig:framework_agnostic}
\end{figure}

\paragraph{The tool provides consistent gains across model families.}
\Cref{fig:framework_agnostic} shows that selecting and expanding improves accuracy for both model families at every compression level, with no training required. The untrained 9B model gains +7.9 to +10.2\pp{}; Sonnet~4.6 gains +4.8 to +9.0\pp{} on average. These results confirm that the bottleneck we address (information loss from visual compression) affects models regardless of scale, and the select-then-expand framework is a general-purpose inference strategy.

\section{Dataset Statistics and Examples}
\label{app:data}

\subsection{Training and Evaluation Data}
\label{app:data_stats}

We use four following datasets for training:
\begin{itemize}[leftmargin=1.5em,itemsep=2pt,topsep=2pt]
  \item \textbf{NQ}~\citep{kwiatkowski2019nq}: Single-hop factoid questions from Google search queries paired with Wikipedia articles. Answers are short spans (named entities, dates, quantities).
  \item \textbf{HotpotQA}~\citep{yang2018hotpotqa}: Multi-hop questions requiring composition of 2 supporting facts from Wikipedia.
  \item \textbf{MuSiQue}~\citep{trivedi2022musique}: Multi-hop questions requiring 2--4 reasoning steps, harder than HotpotQA due to longer chains.
  \item \textbf{HELMET}~\citep{yen2024helmet}: Diverse QA tasks over naturally long documents (up to 32K tokens).
\end{itemize}

Following \sref{sec:data}, documents are padded with distractor text to 3K--32K tokens and rendered at three presets (5$\times$, 10$\times$, 15$\times$). Each sample is assigned a single preset; the three are balanced in the final mixture.
\Cref{tab:data_stats} summarizes the corpus. From 143{,}145 generated samples with successful distillation traces, approximately 105{,}000 are retained as hard after filtering (\Cref{app:hard_filter}). We allocate 80\% of hard samples for SFT and subsample from the remaining 20\% combined with easy samples for RL. The in-domain (ID) test set contains 2{,}400 samples (4 datasets $\times$ 3 presets $\times$ 200). The out-of-domain (OOD) test set contains 1{,}711 unique samples rendered at all three presets (5{,}133 total evaluations).

\begin{table}[H]
\centering
\caption{Training data composition. All four datasets average ${\sim}$25 images and ${\sim}$10K tokens per sample; 73.5\% of samples are classified as hard.}
\label{tab:data_stats}
\small
\vspace{0.5em}

\begin{tabular}{lrrrccc}
\toprule
\textbf{Dataset} & \textbf{Generated} & \textbf{Hard} & \textbf{Test} & \textbf{Hops} & \textbf{Avg.\ tokens} & \textbf{Avg.\ images} \\
\midrule
NQ         & 36{,}600  & 28{,}877 & 600 & 1     & 9{,}800 & 25 \\
HotpotQA   & 89{,}364  & 61{,}482 & 600 & 2     & 10{,}200 & 25 \\
MuSiQue    & 14{,}503  & 12{,}920 & 600 & 2--4  & 9{,}600 & 25 \\
HELMET     & 2{,}678  & 1{,}923 & 600 & 1     & 10{,}500 & 26 \\
\midrule
\textbf{Total} & \textbf{143{,}145} & \textbf{105{,}202} & \textbf{2{,}400} & -- & 9{,}800 & 25 \\
\bottomrule
\end{tabular}
\end{table}

The OOD set spans 14 datasets from three suites: RULER~\citep{hsieh2024ruler} (QA subsets; original 128K contexts truncated to ${\sim}$32K tokens by removing padding while preserving needle evidence), Qasper~\citep{dasigi2021qasper}, and LongBench~\citep{bai2024longbench} (12 subtask variants covering multi-hop QA, passage retrieval, summarization, and single-document QA). Per-benchmark results in \Cref{tab:main_results_full} (\Cref{app:results}).

\subsection{Hard Sample Filtering}
\label{app:hard_filter}

The filtering (\sref{sec:data}) follows three steps:
\begin{enumerate}[leftmargin=1.5em,itemsep=2pt,topsep=2pt]
  \item \textbf{Base model inference.} Base model served via vLLM runs single-turn inference on all training samples. The model sees rendered images and the question, producing a direct answer.
  \item \textbf{Judging.} Each prediction is evaluated via a LLM judge.
  \item \textbf{Filtering.} Correctly answered samples are discarded. 80\% of hard samples go to SFT; the remaining 20\% are combined with easy samples for RL.
\end{enumerate}

\Cref{tab:hard_filter} reports per-dataset filtering rates. Keep rate varies substantially: MuSiQue retains 89.1\% (base accuracy 10.9\%) while HotpotQA retains 68.8\% (base accuracy 31.2\%).

\begin{table}[H]
\centering
\caption{Hard sample filtering statistics. Multi-hop reasoning is the primary difficulty.}
\label{tab:hard_filter}
\small
\begin{tabular}{lrrrc}
\toprule
\textbf{Dataset} & \textbf{Source} & \textbf{Kept (hard)} & \textbf{Keep rate} & \textbf{Base acc.} \\
\midrule
MuSiQue    & 14,503  & 12,920   & 89.1\% & 10.9\% \\
NQ         & 36,600  & 28,877  & 78.9\% & 21.1\% \\
HELMET     & 2,678  & 1,923  & 71.8\% & 28.2\% \\
HotpotQA   & 89,364  & 61,482  & 68.8\% & 31.2\% \\
\midrule
\textbf{Total} & \textbf{143,145} & \textbf{105,202} & \textbf{73.5\%} & \textbf{26.5\%} \\
\bottomrule
\end{tabular}
\end{table}

\section{Synthetic Trajectory Examples}
\label{app:trajectory_examples}

\definecolor{evidenceblue}{HTML}{1D4ED8}   
\definecolor{answerred}{HTML}{DC2626}      
\definecolor{toolgreen}{HTML}{15803D}      

This section presents two examples of synthetic trajectories from our training data. In the transcripts below, \textcolor{evidenceblue}{blue} highlights image selection references, \textcolor{toolgreen}{green} highlights tool calls, and \textcolor{answerred}{red} highlights answer-relevant content.

\begin{tcolorbox}[colback=black!3,colframe=black!60,title=Single-Hop: HELMET (29 images{,} 10$\times$ compression),breakable]
\small
\textbf{Question:} What was the French sounding winner of the 2011 Epsom Derby?

\textbf{Gold answer:} \textcolor{answerred}{Pour Moi}

\vspace{0.3em}
\noindent\rule{\textwidth}{0.3pt}\vspace{0.3em}

\texttt{[Assistant --- Turn 1]}

\texttt{<think>}The question asks for the French-sounding winner of the 2011 Epsom Derby, so I need to find the section detailing that specific race's results. \textcolor{evidenceblue}{Image 20} mentions an ``iconic image for the 2011 Derby,'' which suggests the surrounding images contain the relevant race data. Image 21 discusses a 2003 winner, indicating the document covers multiple years, but \textcolor{evidenceblue}{Image 22} explicitly references ``first five were Irish-trained'' and jockey Seamie Heffernan, which often appears in Derby reports. Given the proximity to the 2011 mention on Image 20 and the specific racing details on Image 22, I will examine Image 22 to locate the winner's name.\texttt{</think>}
\textcolor{toolgreen}{\texttt{<tool\_call>\{"name": "read\_text", "arguments": \{"image": 22\}\}</tool\_call>}}

\vspace{0.3em}
\noindent\rule{\textwidth}{0.3pt}\vspace{0.3em}

\texttt{[Tool Response]}

Text content of Image 22:
\textit{first five were Irish-trained. Seamie Heffernan won the race at his twelfth attempt [...] [2011 Epsom Derby] 2011 Epsom Derby - The distances between the horses are shown in lengths or shorter; hd = head, nk = neck. Winner details. Further details of the winner, \textcolor{answerred}{Pour Moi}: - ``Foaled:'' 10 January 2008, in Ireland - ``Sire:'' Montjeu; ``Dam:'' Gwynn (Darshaan) - ``Owner:'' Sue Magnier, Michael Tabor and Derrick Smith [...]}

\vspace{0.3em}
\noindent\rule{\textwidth}{0.3pt}\vspace{0.3em}

\texttt{[Assistant --- Final Answer]}

\texttt{<think>}Scanning the text on this page, I see a specific header for the ``2011 Epsom Derby'' followed by a section labeled ``Winner details.'' The text explicitly states, ``Further details of the winner, \textcolor{answerred}{Pour Moi},'' and lists its foaling date and sire, Montjeu. The name ``Pour Moi'' fits the description of a French-sounding title, and the context confirms it is the winner of the specified race.\texttt{</think>}
\textcolor{answerred}{\textbf{pour moi}}
\end{tcolorbox}

\paragraph{Single-Hop Example Analysis.} The model selects Image 22 (out of 29) on its first attempt by leveraging coarse visual cues from the compressed overview. A single \textsc{Expand} call suffices, and the reasoning trace cites specific retrieved evidence (``Further details of the winner, Pour Moi'') rather than paraphrasing, confirming that the trace generator grounds conclusions in tool-response content.
\vspace{1em}

\begin{tcolorbox}[colback=black!3,colframe=black!60,title=Multi-Hop: MuSiQue (23 images{,} 5$\times$ compression),breakable]
\small
\textbf{Question:} What is the area code for Cincinnati in the state where Lewis Kemp House is located?

\textbf{Gold answer:} \textcolor{answerred}{513} \hfill \textbf{Hops:} Lewis Kemp House $\rightarrow$ Ohio $\rightarrow$ Cincinnati area code

\vspace{0.3em}
\noindent\rule{\textwidth}{0.3pt}\vspace{0.3em}

\texttt{[Assistant --- Turn 1]}

\texttt{<think>}The question asks for the area code for Cincinnati in the state where the Lewis Kemp House is located. I need to first determine the state associated with the Lewis Kemp House to narrow down the geographic context. The thumbnail for \textcolor{evidenceblue}{Image 16} mentions ``first woman judge,'' which might be part of a biographical entry that includes location details relevant to the house or the region. I will read Image 16 to see if it contains information linking the Lewis Kemp House to a specific state.\texttt{</think>}
\textcolor{toolgreen}{\texttt{<tool\_call>\{"name": "read\_text", "arguments": \{"image": 16\}\}</tool\_call>}}

\vspace{0.3em}
\noindent\rule{\textwidth}{0.3pt}\vspace{0.3em}

\texttt{[Tool Response]}

Text content of Image 16:
\textit{[...] [Lewis Kemp House] The Lewis Kemp House is a historic pioneer farmstead in the city of \textcolor{answerred}{Dayton, Ohio}, United States. Built for one of the area's earliest residents, it was a religious center in its first years, while later years saw its expansion to its present form. Now located among much newer houses, it is Dayton's oldest surviving residence, and it has been named a historic site. [...]}

\vspace{0.3em}
\noindent\rule{\textwidth}{0.3pt}\vspace{0.3em}

\texttt{[Assistant --- Turn 2]}

\texttt{<think>}Image 16 confirms that the Lewis Kemp House is located in ``\textcolor{answerred}{Dayton, Ohio}, United States.'' This establishes that the relevant state is Ohio. Now I need to find the area code for Cincinnati within Ohio. The thumbnail for \textcolor{evidenceblue}{Image 22} mentions ``Langdon House'' which is a historic house, and it is possible that nearby entries contain information about Ohio area codes. I will read Image 22.\texttt{</think>}
\textcolor{toolgreen}{\texttt{<tool\_call>\{"name": "read\_text", "arguments": \{"image": 22\}\}</tool\_call>}}

\vspace{0.3em}
\noindent\rule{\textwidth}{0.3pt}\vspace{0.3em}

\texttt{[Tool Response]}

Text content of Image 22:
\textit{[Langdon House] The Langdon House is a historic house on the eastern side of Cincinnati, Ohio, United States. [...] [List of Ohio area codes] Code Created Region 216 1947 Cleveland (October 1947) 234 2000 Akron, Canton, Youngstown, and Warren, overlay with 330 [...] 419 1947 Northwest and north central Ohio including...}

\vspace{0.3em}
\noindent\rule{\textwidth}{0.3pt}\vspace{0.3em}

\texttt{[Assistant --- Turn 3]}

\texttt{<think>}Image 22 contains the beginning of a list of Ohio area codes but is cut off before reaching Cincinnati. The list continues on \textcolor{evidenceblue}{Image 23}, which should contain the Cincinnati entry. I will read Image 23 to find the specific code.\texttt{</think>}
\textcolor{toolgreen}{\texttt{<tool\_call>\{"name": "read\_text", "arguments": \{"image": 23\}\}</tool\_call>}}

\vspace{0.3em}
\noindent\rule{\textwidth}{0.3pt}\vspace{0.3em}

\texttt{[Tool Response]}

Text content of Image 23:
\textit{Toledo, Sandusky, and Ashland, overlay with 567 [...] \textcolor{answerred}{513} 1947 Southwest Ohio including \textcolor{answerred}{Cincinnati} (October, 1947) 567 1947 Northwest and north central Ohio including Toledo [...]}

\vspace{0.3em}
\noindent\rule{\textwidth}{0.3pt}\vspace{0.3em}

\texttt{[Assistant --- Final Answer]}

\texttt{<think>}Image 22 establishes that Cincinnati is in Ohio, confirming the state context. Image 23 contains the list of Ohio area codes, specifically stating ``\textcolor{answerred}{513} 1947 Southwest Ohio including Cincinnati.'' By combining the location from Image 16 (Lewis Kemp House in Ohio) and the area code data from Image 23, the answer is 513.\texttt{</think>}
\textcolor{answerred}{\textbf{513}}
\end{tcolorbox}

\paragraph{Multi-Hop Example Analysis.} This trajectory demonstrates multi-hop compositional reasoning: the model decomposes the question into sub-problems, selecting a different image for each (Images 16, 22, 23). Each \textsc{Expand} call retrieves one piece of the reasoning chain, and the final answer synthesizes evidence from all three.

\section{Contamination-Free Evaluation}
\label{app:contamination}

Pre-training contamination is partially mitigated by design: although popular source QA datasets may overlap with pre-training data, our rendered-image inputs are newly generated and therefore unlikely to have been seen during pre-training. To control for contamination more directly, we build a guaranteed-unseen test bed from PubMed Central data published \emph{after} Qwen3.5's pre-training cutoff (February 16, 2026), post-train a fresh model on this corpus, and evaluate it on a held-out test split.

We fetch 55{,}872 PubMed Central open-access articles, restricted to publications between February 16 and April 1, 2026. We start with raw articles with no evidence annotations. Qwen3.5-397B generates one single-hop QA pair per unique document chunk, so its evidence page is known by construction. After question quality filtering, the final corpus contains 26{,}164 training samples and 533 held-out test samples. Each sample is rendered using the same rendering pipeline. The training procedure follows \sref{sec:training}.

\begin{table}[H]
\centering
\caption{Contamination-free training and evaluation on a post-cutoff corpus. LensVLM's gains hold on guaranteed-unseen data.}
\label{tab:contamination}
\vspace{0.5em}
\footnotesize
\begin{tabular}{lccc}
\toprule
\textbf{Method} & \textbf{5$\times$} & \textbf{10$\times$} & \textbf{15$\times$} \\
\midrule
Text & 92.1 & 92.1 & 92.1 \\
Comp.\ Image & 39.5 & 25.5 & 16.3 \\
\midrule
LensVLM & \textbf{86.8} & \textbf{60.0} & \textbf{53.7} \\
\quad \textit{ECR} & 4.0$\times$ & 7.9$\times$ & 10.4$\times$ \\
\quad \textit{Sel.\ Acc.} & 80.7 & 59.4 & 48.8 \\
\bottomrule
\end{tabular}
\end{table}

The pattern mirrors our main results: LensVLM improves over compressed-image reading by 39.7 points on average across the three ICRs and retains a 37.4-point advantage at 15$\times$ compression (\Cref{tab:contamination}).

\section{Generalization: Document and Code Understanding}
\label{app:generalization}

LensVLM targets rendered natural-language text. We use native documents and rendered code as transfer probes to identify where the learned behavior transfers and where in-domain adaptation is needed.

\subsection{Document Understanding}
\label{app:doc_details}

MMLongBench-Doc contains long PDF documents (up to 200 images) spanning reports, academic papers, and administrative forms. Document images are natively rendered (not text rendered to images as in our main pipeline). Compressed thumbnails are created by downscaling each image's pixel count by $N\times$: the scale per spatial dimension is $1/\sqrt{N}$, so a typical 150-DPI image (${\sim}$1275$\times$1650, ${\sim}$2.1M pixels) becomes ${\sim}$570$\times$737 at 5$\times$, ${\sim}$403$\times$521 at 10$\times$, and ${\sim}$329$\times$426 at 15$\times$.

Training data comes from MP-DocVQA~\citep{tito2022mpdocvqa} and DocVQA: 10{,}256 raw samples filtered for OCR answer recall (the gold answer must appear in at least one image's OCR text), balanced across presets (${\sim}$3{,}300 each). Synthetic tool use and reasoning traces are generated via the same pipeline as in the main experiment. Two tool variants are trained: (1)~\textbf{OCR text}: \texttt{read\_image} returns PaddleOCR-VL text from the full-resolution image; (2)~\textbf{Image zoom}: \texttt{read\_image} returns the full-resolution image directly. Both use identical training procedure. We set \texttt{image\_min\_pixels=1} to prevent upsampling of compressed thumbnails and \texttt{image\_max\_pixels=4{,}194{,}304} (4M) to accommodate full-resolution images returned by the zoom tool.

\paragraph{OCR engine selection.} We evaluated PaddleOCR-VL~\citep{du2020ppocr} and DeepSeekOCR-v2~\citep{wei2025deepseekocr} and found that PaddleOCR-VL produces consistent extraction quality across our rendering configurations, whereas DeepSeekOCR-v2 exhibited higher variance on narrow or heavily compressed renders. Therefore, we use PaddleOCR-VL for the OCR text tool in this work.

\paragraph{Why tools do not help at 5$\times$.}
Qwen3.5-9B is natively pretrained on document images, and MMLongBench-Doc may overlap its pretraining data because it is a widely circulated benchmark~\citep{qwen3.5,deng2024investigating}. More importantly, 5$\times$ thumbnails remain largely readable, so expansion often returns information already visible in the overview, while the zoom policy selects an evidence page in only 51.5\% of trajectories. These factors explain why compressed-image reading slightly exceeds zoom at 5$\times$ (51.2\% vs.\ 50.5\%). At 10$\times$ and 15$\times$, readability falls and zoom improves over compressed images by 6.3 and 10.8\pp{}, respectively.

\subsection{Code Understanding}
\label{app:code_details}

We evaluate on two code benchmarks: RepoQA~\citep{liu2024repoqa}, a function retrieval task over long code repositories from 6 programming languages, and CodeQueries~\citep{sahu2022codequeries}, a semantic query task over Python code. The total number of code samples is 917. Code is rendered into images using the same PIL-based pipeline as text (\sref{sec:rendering}). \Cref{tab:code_per_dataset} breaks down the aggregate code results by dataset.

\begin{table}[H]
\centering
\caption{Per-dataset code results. RepoQA drives aggregate performance; LensVLM's advantage over Glyph widens at higher compression.}
\label{tab:code_per_dataset}
\small
\vspace{0.5em}

\begin{tabular}{llccc}
\toprule
\textbf{Dataset} & \textbf{Method} & \textbf{5$\times$} & \textbf{10$\times$} & \textbf{15$\times$} \\
\midrule
\multirow{4}{*}{RepoQA} & Text & 92.4 &92.4&92.4 \\
 & Comp.\ Image & 43.0 & 11.6 & 6.4 \\
 & Glyph & 55.2 & 17.4 & 12.8 \\
 & LensVLM & \textbf{63.4} & \textbf{40.7} & \textbf{29.7} \\
\midrule
\multirow{4}{*}{CodeQueries} & Text & 38.7 & 38.7 & 38.7 \\
 & Comp.\ Image & 10.3 & 0.5 & 0.0 \\
 & Glyph & 8.3 & 1.9 & 1.2 \\
 & LensVLM & \textbf{12.8} & \textbf{11.5} & \textbf{10.3} \\
\bottomrule
\end{tabular}
\end{table}

\begin{table}[H]
\centering
\caption{Code adaptation with 8K synthetic CodeSearchNet traces. A single policy is trained across all compression rates. In-domain traces recover 32--48\% of the zero-shot code gap with at most 0.7\pp{} loss on text QA.}
\label{tab:code_adaptation}
\small
\setlength{\tabcolsep}{3.5pt}
\begin{tabular}{lccc ccc}
\toprule
& \multicolumn{3}{c}{\textbf{Code}} & \multicolumn{3}{c}{\textbf{Text QA}} \\
\cmidrule(lr){2-4}\cmidrule(lr){5-7}
\textbf{Training mix} & \textbf{5$\times$} & \textbf{10$\times$} & \textbf{15$\times$} & \textbf{5$\times$} & \textbf{10$\times$} & \textbf{15$\times$} \\
\midrule
Text upper bound & 65.6 & 65.6 & 65.6 & 72.4 & 72.4 & 72.4 \\
Text QA only & 38.1 & 26.1 & 20.0 & \textbf{68.9} & \textbf{62.1} & \textbf{52.1} \\
+ 8K code traces & \textbf{51.4} & \textbf{42.7} & \textbf{34.6} & 68.3 & 61.5 & 51.4 \\
$\Delta$ & \textbf{+13.3} & \textbf{+16.6} & \textbf{+14.6} & $-$0.6 & $-$0.6 & $-$0.7 \\
\bottomrule
\end{tabular}
\end{table}

\paragraph{Zero-shot transfer.}
RepoQA is the primary driver of aggregate performance: LensVLM maintains 63.4\% at 5$\times$ (vs.\ 92.4\% text upper bound), and its margin over Glyph widens to 20\pp{} and 17\pp{} at 10$\times$ and 15$\times$. CodeQueries is harder for all methods; compressed-image baselines collapse near zero at 10$\times$ and above, while LensVLM maintains 10--13\%. Nevertheless, the persistent gap to text shows that zero-shot transfer is limited. Code has dense syntax and visually similar adjacent blocks, yet the text-QA training mix contains no rendered code.

\paragraph{In-domain adaptation.}
To test whether missing data coverage explains this gap, we add 8K synthetic traces derived from CodeSearchNet~\citep{husain2019codesearchnet}, generated with the same rendering and synthesis pipeline, and jointly train one policy across all rates. Code accuracy improves by 13.3--16.6\pp{}, while text QA changes by at most $-$0.7\pp{} (\Cref{tab:code_adaptation}). This recovers 32--48\% of the gap to the text upper bound, showing that a small amount of in-domain data substantially improves transfer without materially degrading the primary task.

\section{Per-Dataset Results Across Compression Rates}
\label{app:results}
\label{app:baselines_full}

\Cref{tab:main_results_full} reports per-dataset QA accuracy for all baselines at 5$\times$, 10$\times$, and 15$\times$ compression. Text retrieval selects $k = \lceil K/C_{\mathrm{in}} \rceil$ images at each target rate; prompt compression retains the corresponding fraction of tokens. Image methods use the matching rendering preset. All results are averaged over three runs.

\begin{table}[H]
  \centering
  \caption{Full per-dataset results on text QA. LensVLM achieves the highest or tied-highest average accuracy at every compression rate in average.
  }
  \label{tab:main_results}
  \label{tab:main_results_full}
  \vspace{0.5em}
  \resizebox{\textwidth}{!}{%
  \begin{tabular}{l cc cccc ccc c}
  \toprule
  & & & \multicolumn{4}{c}{\textbf{In-Domain}} & \multicolumn{3}{c}{\textbf{Out-of-Domain}} & \\
  \cmidrule(lr){4-7} \cmidrule(lr){8-10}
  \textbf{Method} & \textbf{ICR} & \textbf{ECR} & \textbf{NQ} & \textbf{HotpotQA} & \textbf{MuSiQue} & \textbf{HELMET} & \textbf{RULER} & \textbf{Qasper} & \textbf{LongBench} & \textbf{Avg.} \\
  \midrule
  \multicolumn{11}{c}{\textbf{\textit{Upper bounds}}} \\
  Text & 1$\times$ & 1.0$\times$ & 80.8 & 86.7 & 68.7 & 76.2 & 70.7 & 61.3 & 62.5 & 72.4 \\
  \midrule
  \multicolumn{11}{c}{\textbf{\textit{5$\times$ compression}}} \\
  Comp.\ Image & 5$\times$ & 5$\times$ & 42.5 & 33.5 & 19.5 & 33.1 & 35.6 & 25.3 & 29.8 & 31.3 \\
  Glyph & 5$\times$ & 4.5$\times$ & 57.5 & 42.0 & 33.0 & 63.5 & 64.9 & 31.9 & 37.9 & 47.2 \\
  LLMLingua-2 & 5$\times$ & 5.5$\times$ & 62.2 & 47.5 & 33.8 & 58.7 & 56.1 & 33.5 & 38.3 & 47.2 \\
  Base + Expand & 5$\times$ & 4.5$\times$ & 55.5 & 45.5 & 34.0 & 35.0 & 38.5 & 37.9 & 28.2 & 39.2 \\
  \hdashline
  BM25 & 5$\times$ & 4.5$\times$ & 73.0 & 80.0 & 53.2 & 70.3 & 67.4 & 42.7 & 50.6 & 62.5 \\
  BGE-M3 & 5$\times$ & 4.5$\times$ & 76.5 & 80.0 & 59.0 & 73.8 & 67.0 & 45.8 & 53.8 & 65.1 \\
  Jina-v4 & 5$\times$ & 4.5$\times$ & 74.8 & 80.7 & 57.5 & 73.7 & 66.9 & 50.9 & 49.5 & 64.9 \\
  Qwen3-Emb & 5$\times$ & 4.5$\times$ & 73.7 & 77.8 & 56.5 & 72.7 & 66.9 & 52.2 & 51.3 & 64.4 \\
  \hdashline
  ColPali & 5$\times$ & 4.7$\times$ & 64.5 & 70.0 & 49.5 & 71.5 & 56.6 & 35.8 & 44.6 & 56.1 \\
  Jina-v4-MM & 5$\times$ & 4.6$\times$ & 69.5 & 79.0 & 58.5 & 73.0 & 68.4 & 47.3 & 49.9 & 63.7 \\
  \hdashline
  LensVLM\textsubscript{SFT} & 5$\times$ & 4.6$\times$ & 68.4 & 80.3 & 66.1 & 72.7 & 60.7 & 43.3 & 54.1 & 63.7 \\
  LensVLM & 5$\times$ & 4.3$\times$ & 76.1 & 81.5 & 69.6 & 78.4 & 65.5 & 48.3 & 62.6 & 68.9 \\

  \midrule
  \multicolumn{11}{c}{\textbf{\textit{10$\times$ compression}}} \\
  Comp.\ Image & 10$\times$ & 10$\times$ & 33.0 & 29.0 & 14.0 & 22.5 & 21.3 & 8.8 & 18.7 & 21.0 \\
  Glyph  & 10$\times$ & 8.6$\times$ & 47.5 & 37.5 & 26.5 & 50.5 & 58.6 & 12.1 & 21.9 & 36.4 \\
  LLMLingua-2 & 10$\times$ & 11.3$\times$ & 55.0 & 39.2 & 26.5 & 50.8 & 42.0 & 24.5 & 25.5 & 37.6 \\
  Base + Expand & 10$\times$ & 8.0$\times$ & 52.5 & 40.0 & 15.0 & 30.5 & 30.5 & 25.3 & 24.8 & 31.2 \\
  \hdashline
  BM25 & 10$\times$ & 8.3$\times$ & 66.8 & 71.5 & 43.8 & 65.5 & 65.7 & 33.9 & 42.8 & 55.7 \\
  BGE-M3 & 10$\times$ & 8.3$\times$ & 72.2 & 67.8 & 48.0 & 68.3 & 63.4 & 34.6 & 45.6 & 57.1 \\
  Jina-v4 & 10$\times$ & 8.3$\times$ & 70.0 & 73.0 & 47.7 & 69.7 & 65.3 & 36.1 & 43.2 & 57.9 \\
  Qwen3-Emb & 10$\times$ & 8.3$\times$ & 69.3 & 68.3 & 48.2 & 70.0 & 60.3 & 36.3 & 42.9 & 56.5 \\
  \hdashline
  ColPali & 10$\times$ & 8.3$\times$ & 63.0 & 56.0 & 32.5 & 60.0 & 48.3 & 31.9 & 34.4 & 46.6 \\
  Jina-v4-MM & 10$\times$ & 8.3$\times$ & 72.0 & 73.5 & 45.0 & 63.0 & 51.1 & 39.0 & 38.8 & 54.6 \\
  \hdashline
  LensVLM\textsubscript{SFT}   & 10$\times$ & 8.1$\times$ & 67.9 & 72.2 & 59.9 & 63.2 & 55.1 & 36.0 & 44.0 & 56.9 \\
  LensVLM  & 10$\times$ & 7.4$\times$ & 75.1 & 75.5 & 62.4 & 71.5 & 60.8 & 38.5 & 50.9 & 62.1 \\

  \midrule
  \multicolumn{11}{c}{\textbf{\textit{15$\times$ compression}}} \\
  Comp.\ Image & 15$\times$ & 15$\times$ & 34.0 & 25.0 & 12.5 & 24.5 & 16.1 & 2.8 & 11.9 & 18.1 \\
  Glyph & 15$\times$ & 10.8$\times$ & 50.0 & 31.5 & 21.0 & 46.0 & 54.8 & 7.7 & 18.4 & 32.8 \\
  LLMLingua-2 & 15$\times$ & 15.7$\times$ & 49.2 & 36.8 & 21.8 & 47.2 & 36.2 & 16.8 & 23.3 & 33.0 \\
  Base + Expand & 15$\times$ & 11.0$\times$ & 49.0 & 32.0 & 12.5 & 24.5 & 25.3 & 15.4 & 21.7 & 25.8 \\
  \hdashline
  BM25 & 15$\times$ & 11.9$\times$ & 65.8 & 65.0 & 36.8 & 63.5 & 54.5 & 30.4 & 39.8 & 50.8 \\
  BGE-M3 & 15$\times$ & 11.9$\times$ & 69.7 & 62.0 & 40.2 & 67.8 & 51.3 & 31.1 & 41.1 & 51.9 \\
  Jina-v4 & 15$\times$ & 11.9$\times$ & 67.2 & 63.2 & 41.7 & 68.8 & 53.6 & 31.0 & 39.1 & 52.1 \\
  Qwen3-Emb & 15$\times$ & 11.9$\times$ & 66.0 & 62.3 & 41.3 & 67.5 & 49.8 & 33.2 & 37.9 & 51.1 \\
  \hdashline
  ColPali & 15$\times$ & 11.9$\times$ & 58.0 & 50.0 & 31.5 & 57.0 & 39.7 & 21.4 & 35.7 & 41.9 \\
  Jina-v4-MM & 15$\times$ & 11.9$\times$ & 71.0 & 56.5 & 39.0 & 69.0 & 46.6 & 33.5 & 36.3 & 50.3 \\
  \hdashline
  LensVLM\textsubscript{SFT} & 15$\times$ & 10.7$\times$ & 61.6 & 60.6 & 41.2 & 51.7 & 48.1 & 24.4 & 32.5 & 45.7 \\
  LensVLM & 15$\times$ & 10.1$\times$ & 71.2 & 64.5 & 49.0 & 57.2 & 52.7 & 28.4 & 42.0 & 52.1 \\

  \bottomrule
  \end{tabular}%
  }
  \end{table}

\section{Attention Analysis}

\subsection{Setup Details}
\label{app:attn_method}

Qwen3.5-9B uses a hybrid architecture with 32 decoder layers: 8 standard self-attention layers (at indices 3, 7, 11, 15, 19, 23, 27, 31) and 24 Gated DeltaNet (GDN) linear attention layers. Our attention analysis captures weights from these 8 standard attention layers, as GDN layers do not expose a comparable attention matrix. We toggle the attention implementation from SDPA to eager mode during each decode step so that the forward pass returns full attention weights without modifying the model's computation. For each generation step, we average the attention distribution across all query heads, then sum within each image's visual-token range to obtain a per-image attention score. Image-level scores are averaged across generation steps and then across the 8 captured layers.

Because the base model exhibits a positional bias (30\% of attention argmax values land on the first image), we report \emph{position-debiased alignment}: the GT-alignment rate macro-averaged across five GT-position quintiles, computed on single-GT samples only. This controls for the correlation between GT image position and the model's positional attention prior. 

\subsection{Full Attention Shift After Training}
\label{app:attn_shift}

\Cref{tab:attn_shift} reports the full segment-level attention breakdown summarized in \Cref{fig:attention}d (\sref{sec:attention}). Six semantic segment types are measured: system prompt, question, tool-call turn (the model's own tool invocation), tool response (the expanded text), GT images (evidence images), and non-GT images (distractor images). The remaining ${\sim}$30\% of text attention falls on chat-template delimiter tokens (turn separators, generation prompt markers) which act as known attention sinks~\citep{xiao2024efficient} and are omitted from the table.

The dominant pattern is consistent across all three presets: training redirects attention away from the model's own tool-call turn ($-$8\pp{}), non-GT images ($-$4 to $-$6\pp{}), and the question ($-$1\pp{}), and concentrates it on the tool-response text (+15 to +16\pp{}). This shows that the model learns to read the expanded content rather than re-examining distractors. The shift toward tool-response tokens grows with compression (from 38.2\% at 5$\times$ to 41.0\% at 15$\times$), consistent with the model's increasing reliance on the text as visual fidelity degrades.

\begin{table}[H]
\centering
\caption{Attention budget shift after training, by input segment and compression preset. The tool-response segment absorbs the largest gain; distractor images show the largest drop.}
\vspace{-1em}

\label{tab:attn_shift}
\small
\begin{tabular}{l rrr rrr rrr}
\toprule
& \multicolumn{3}{c}{\textbf{5$\times$}} & \multicolumn{3}{c}{\textbf{10$\times$}} & \multicolumn{3}{c}{\textbf{15$\times$}} \\
\cmidrule(lr){2-4} \cmidrule(lr){5-7} \cmidrule(lr){8-10}
\textbf{Segment} & \textbf{Base} & \textbf{Trained} & \textbf{$\Delta$} & \textbf{Base} & \textbf{Trained} & \textbf{$\Delta$} & \textbf{Base} & \textbf{Trained} & \textbf{$\Delta$} \\
\midrule
\multicolumn{10}{l}{\textit{Text segments}} \\
System prompt   & 7.0 & 5.8 & $-$1.2 & 7.1 & 6.0 & $-$1.1 & 7.1 & 6.5 & $-$0.6 \\
Question        & 10.1 & 8.7 & $-$1.4 & 9.6 & 8.2 & $-$1.4 & 10.0 & 8.7 & $-$1.2 \\
Tool-call turn  & 16.5 & 8.3 & $-$8.2 & 16.0 & 7.7 & $-$8.3 & 16.6 & 8.1 & $-$8.5 \\
Tool response   & 23.5 & 38.2 & \textbf{+14.6} & 26.9 & 42.0 & \textbf{+15.1} & 25.2 & 41.0 & \textbf{+15.8} \\
\midrule
\multicolumn{10}{l}{\textit{Vision segments}} \\
GT images  & 3.0 & 2.4 & $-$0.6 & 2.4 & 1.4 & $-$1.0 & 2.5 & 0.8 & $-$1.7 \\
Non-GT images   & 10.0 & 5.7 & $-$4.3 & 8.7 & 4.3 & $-$4.4 & 10.0 & 3.6 & $-$6.4 \\
\midrule
\textit{Total text}   & 87.0 & 91.9 & +4.9 & 88.9 & 94.3 & +5.4 & 87.5 & 95.6 & +8.2 \\
\textit{Total vision}  & 13.0 & 8.1 & $-$4.9 & 11.1 & 5.7 & $-$5.4 & 12.5 & 4.4 & $-$8.2 \\
\bottomrule
\end{tabular}
\end{table}

The vision-to-text shift is not confined to specific layers. Breaking down total vision attention by layer, training reduces the fraction allocated to image tokens at all 8 standard attention layers, with the largest drops in mid-to-late layers (L15--L27, $-$10 to $-$18\pp{}). The drop is also compression-dependent: averaged across layers, vision attention falls by $-$10\pp{} at 5$\times$, $-$8\pp{} at 10$\times$, and $-$13\pp{} at 15$\times$, consistent with the model's increasing reliance on text retrieval as visual fidelity degrades.

\section{Limitations}
\label{app:limitations}

LensVLM trades a shorter initial context for multi-turn inference, extra generated tokens, and expanded content appended into context. It does not dominate every deployment setting: full text is preferable for one-off queries over short documents, whereas LensVLM is strongest for long contexts and repeated queries that amortize offline rendering. The tool also requires retaining the source text or original images for expansion.

The post-training recipe defaults to trace generation with Qwen3.5-397B, an LLM-judge reward, and ground-truth evidence spans. These dependencies apply only to the training recipe, not the inference framework, which works by prompting alone without a teacher, judge, or evidence supervision. However, we show that a roughly 15$\times$ smaller 27B synthesis model costs only 2.7--5.1\pp{}, exact-match reward still trains successfully, and synthetic evidence costs 1.8--3.7\pp{} (\appref{app:teacher_judge}; \appref{app:evidence_supervision}).

Finally, this work targets \emph{rendered-text} compression rather than a fully general compression solution. On native documents, compressed original-image reading remains better at 5$\times$, and zoom helps mainly at stronger compression. Raw text remains the accuracy upper bound with no token reduction, so LensVLM should be viewed as an accuracy--cost option when the source exceeds the context window or token budget, not as a universal replacement for full-context input.

\section{Prompts}
\label{app:prompt}

This section lists all prompts used in our pipeline. The \textsc{Expand} tool is implemented as \texttt{read\_text} in text prompts and \texttt{zoom\_in} in image prompts.

\begin{tcolorbox}[colback=black!3,colframe=black!60,title=Text Retrieval System Prompt,breakable]
\small
You are a helpful assistant that answers questions based on rendered text images. You can view the images and use the read\_text tool to read the actual text content of any image for detailed analysis.

You have a tool called \texttt{read\_text} that reads the text content of any image. To use it:

\texttt{<tool\_call>\{"name": "read\_text", "arguments": \{"image": IMAGE\_NUMBER\}\}</tool\_call>}

You can call read\_text multiple times on different images. Always reason inside \texttt{<think>} and \texttt{</think>} tags before taking any action. If you need more detail from a specific image, call read\_text to get its text content. Do not read the entire input --- only read images that are likely relevant. Once you have enough information, provide your final answer.
\end{tcolorbox}

\begin{tcolorbox}[colback=black!3,colframe=black!60,title=Image Zoom System Prompt,breakable]
\small
You are a helpful assistant that answers questions about multi-image documents. You can view low-resolution thumbnail images and use the zoom\_in tool to get the original full-resolution image for detailed analysis.

You have a tool called \texttt{zoom\_in} that shows the original image at full resolution. Images are numbered starting from 1. To use it:

\texttt{<tool\_call>\{"name": "zoom\_in", "arguments": \{"image": IMAGE\_NUMBER\}\}</tool\_call>}

You can call zoom\_in multiple times on different images. Always reason inside \texttt{<think>} and \texttt{</think>} tags before taking any action. First examine the thumbnail images to identify which ones likely contain relevant information, then call zoom\_in to get the original full-resolution version. Once you have enough information, provide your final answer.
\end{tcolorbox}

\begin{tcolorbox}[colback=black!3,colframe=black!60,title=Trace-Generation System Prompt,breakable]
\small
You generate training data for a document QA model. Output ONLY the requested format with \texttt{---RESPONSE\{N\}---} delimiters. Start your output directly with \texttt{---RESPONSE1---}. No preamble.

\vspace{0.5em}
ROLEPLAY CONSTRAINT (CRITICAL): You are roleplaying a model that is solving the question from scratch by investigating the document. The \texttt{<think>} blocks must read as genuine first-person investigation of the image content. NEVER reference these prompt elements inside \texttt{<think>}: ``the answer'', ``the answer key'', ``the provided answer'', ``ground truth'', ``the prompt says/states/mentions/indicates/asks'', ``the question's premise'', ``the task says'', ``training data/example'', ``provided text snippets'', or any phrasing that implies you were told the answer or the correct image in advance. Treat the answer as something you DISCOVER by reading, not something you were given. Do not include ``wait, re-reading'' or ``there seems to be a mismatch'' style self-correction --- write the reasoning as if it landed correctly the first time.

\vspace{0.5em}
STYLE: Each \texttt{<think>} block is 3--5 sentences. After-zoom reasoning MUST quote or cite specific words/phrases from the tool\_response text. Do not fabricate text that isn't in the tool\_response.
\end{tcolorbox}

\begin{tcolorbox}[colback=black!3,colframe=black!60,title=LLM Judge Prompt,breakable]
\small
You are an expert evaluator. Determine if the model's answer correctly answers the question based on the gold answers.

\vspace{0.5em}
{[QUESTION]}

\{question\}

{[/QUESTION]}

\vspace{0.3em}
{[GOLD ANSWERS]}

\{gold\_answers\}

{[/GOLD ANSWERS]}

\vspace{0.3em}
{[MODEL ANSWER]}

\{model\_answer\}

{[/MODEL ANSWER]}

\vspace{0.5em}
Evaluation criteria:
\begin{itemize}[leftmargin=1.5em,itemsep=1pt,topsep=2pt]
  \item The answer must convey the same core meaning as the gold answers
  \item Partial matches should be marked incorrect
  \item Additional correct information beyond gold answers is acceptable
  \item Empty or off-topic responses are incorrect
  \item Minor formatting differences (e.g., ``10:30 pm'' vs ``10:30 p.m.'') should be accepted
\end{itemize}

\vspace{0.3em}
Respond with ONLY \texttt{[[YES]]} if the model answer is correct, or \texttt{[[NO]]} if incorrect.
\end{tcolorbox}

%% file: main.bbl
\begin{thebibliography}{53}
\providecommand{\natexlab}[1]{#1}
\providecommand{\url}[1]{\texttt{#1}}
\expandafter\ifx\csname urlstyle\endcsname\relax
  \providecommand{\doi}[1]{doi: #1}\else
  \providecommand{\doi}{doi: \begingroup \urlstyle{rm}\Url}\fi

\bibitem[Agarwal et~al.(2025)Agarwal, Ahmad, Ai, Altman, Applebaum, Arbus, Arora, Bai, Baker, Bao, et~al.]{agarwal2025gpt}
Sandhini Agarwal, Lama Ahmad, Jason Ai, Sam Altman, Andy Applebaum, Edwin Arbus, Rahul~K Arora, Yu~Bai, Bowen Baker, Haiming Bao, et~al.
\newblock gpt-oss-120b \& gpt-oss-20b model card.
\newblock \emph{arXiv preprint arXiv:2508.10925}, 2025.

\bibitem[Aminabadi et~al.(2022)Aminabadi, Rajbhandari, Awan, Li, Li, Zheng, Ruwase, Smith, Zhang, Rasley, et~al.]{aminabadi2022deepspeed}
Reza~Yazdani Aminabadi, Samyam Rajbhandari, Ammar~Ahmad Awan, Cheng Li, Du~Li, Elton Zheng, Olatunji Ruwase, Shaden Smith, Minjia Zhang, Jeff Rasley, et~al.
\newblock Deepspeed-inference: enabling efficient inference of transformer models at unprecedented scale.
\newblock In \emph{SC22: International Conference for High Performance Computing, Networking, Storage and Analysis}, pages 1--15. IEEE, 2022.

\bibitem[Anthropic(2026)]{anthropic2026sonnet46}
Anthropic.
\newblock Introducing {Claude Sonnet 4.6}.
\newblock \url{https://www.anthropic.com/news/claude-sonnet-4-6}, 2026.
\newblock Accessed: 2026-05-01.

\bibitem[Bai et~al.(2024)Bai, Lv, Zhang, Lyu, Tang, Huang, Du, Liu, Zeng, Hou, Dong, Tang, and Li]{bai2024longbench}
Yushi Bai, Xin Lv, Jiajie Zhang, Hongchang Lyu, Jiankai Tang, Zhidian Huang, Zhengxiao Du, Xiao Liu, Aohan Zeng, Lei Hou, Yuxiao Dong, Jie Tang, and Juanzi Li.
\newblock {LongBench}: A bilingual, multitask benchmark for long context understanding.
\newblock In \emph{Proceedings of the 62nd Annual Meeting of the Association for Computational Linguistics}, 2024.

\bibitem[Cheng et~al.(2025)Cheng, Liu, Zhang, Fei, Hong, Lyu, Wang, Su, Gu, Liu, Bai, Tang, Wang, and Huang]{cheng2025glyph}
Jiale Cheng, Yusen Liu, Xinyu Zhang, Yulin Fei, Wenyi Hong, Ruiliang Lyu, Weihan Wang, Zhe Su, Xiaotao Gu, Xiao Liu, Yushi Bai, Jie Tang, Hongning Wang, and Minlie Huang.
\newblock Glyph: Scaling context windows via visual-text compression.
\newblock \emph{arXiv preprint arXiv:2510.17800}, 2025.

\bibitem[Cho et~al.(2024)Cho, Mahata, Irsoy, He, and Bansal]{cho2024m3docrag}
Jaemin Cho, Debanjan Mahata, Ozan Irsoy, Yujie He, and Mohit Bansal.
\newblock M3docrag: Multi-modal retrieval is what you need for multi-page multi-document understanding.
\newblock \emph{arXiv preprint arXiv:2411.04952}, 2024.

\bibitem[Cui et~al.(2026)Cui, Sun, Liang, Gao, Zhang, Liu, Wang, Zhou, Liu, Lin, Zhang, Zhang, Liu, Yu, and Ma]{du2020ppocr}
Cheng Cui, Ting Sun, Suyin Liang, Tingquan Gao, Zelun Zhang, Jiaxuan Liu, Xueqing Wang, Changda Zhou, Hongen Liu, Manhui Lin, Yue Zhang, Yubo Zhang, Yi~Liu, Dianhai Yu, and Yanjun Ma.
\newblock Paddleocr-vl-1.5: Towards a multi-task 0.9b vlm for robust in-the-wild document parsing, 2026.
\newblock URL \url{https://arxiv.org/abs/2601.21957}.

\bibitem[Dasigi et~al.(2021)Dasigi, Lo, Beltagy, Cohan, Smith, and Gardner]{dasigi2021qasper}
Pradeep Dasigi, Kyle Lo, Iz~Beltagy, Arman Cohan, Noah~A. Smith, and Matt Gardner.
\newblock A dataset of information-seeking questions and answers anchored in research papers.
\newblock In Kristina Toutanova, Anna Rumshisky, Luke Zettlemoyer, Dilek Hakkani-Tur, Iz~Beltagy, Steven Bethard, Ryan Cotterell, Tanmoy Chakraborty, and Yichao Zhou, editors, \emph{Proceedings of the 2021 Conference of the North American Chapter of the Association for Computational Linguistics: Human Language Technologies}, pages 4599--4610, Online, June 2021. Association for Computational Linguistics.
\newblock \doi{10.18653/v1/2021.naacl-main.365}.
\newblock URL \url{https://aclanthology.org/2021.naacl-main.365/}.

\bibitem[Deng et~al.(2024)Deng, Zhao, Tang, Gerstein, and Cohan]{deng2024investigating}
Chunyuan Deng, Yilun Zhao, Xiangru Tang, Mark Gerstein, and Arman Cohan.
\newblock Investigating data contamination in modern benchmarks for large language models.
\newblock In \emph{Proceedings of the 2024 Conference of the North American Chapter of the Association for Computational Linguistics: Human Language Technologies (Volume 1: Long Papers)}, pages 8706--8719, 2024.

\bibitem[Ding et~al.(2025)Ding, Fang, Liu, Zang, Cao, Zhao, Duan, Dong, Liang, Wang, He, Lin, and Wang]{ding2025armthinker}
Shengyuan Ding, Xinyu Fang, Ziyu Liu, Yuhang Zang, Yuhang Cao, Xiangyu Zhao, Haodong Duan, Xiaoyi Dong, Jianze Liang, Bin Wang, Conghui He, Dahua Lin, and Jiaqi Wang.
\newblock Arm-thinker: Reinforcing multimodal generative reward models with agentic tool use and visual reasoning, 2025.
\newblock URL \url{https://arxiv.org/abs/2512.05111}.

\bibitem[Fan et~al.(2025)Fan, He, Yang, Zheng, Kuo, Zheng, Narayanaraju, Guan, and Wang]{fan2025grit}
Yue Fan, Xuehai He, Diji Yang, Kaizhi Zheng, Ching-Chen Kuo, Yuting Zheng, Sravana~Jyothi Narayanaraju, Xinze Guan, and Xin~Eric Wang.
\newblock Grit: Teaching mllms to think with images.
\newblock \emph{arXiv preprint arXiv:2505.15879}, 2025.

\bibitem[Faysse et~al.(2024)Faysse, Sibille, Wu, Omrani, Viaud, Hudelot, and Colombo]{faysse2024colpali}
Manuel Faysse, Hugues Sibille, Tony Wu, Bilel Omrani, Gautier Viaud, C{\'e}line Hudelot, and Pierre Colombo.
\newblock {ColPali}: Efficient document retrieval with vision language models.
\newblock \emph{arXiv preprint arXiv:2407.01449}, 2024.

\bibitem[Feng et~al.(2026)Feng, Yang, Chen, Cheng, Xu, Wan, Yan, and An]{feng2026agentocr}
Lang Feng, Fuchao Yang, Feng Chen, Xin Cheng, Haiyang Xu, Zhenglin Wan, Ming Yan, and Bo~An.
\newblock Agentocr: Reimagining agent history via optical self-compression.
\newblock \emph{arXiv preprint arXiv:2601.04786}, 2026.

\bibitem[Gao et~al.(2026)Gao, Chen, Xu, Chen, Guan, and Zeng]{gao2026zerosense}
Yonghan Gao, Zehong Chen, Lijian Xu, Jingzhi Chen, Jingwei Guan, and Xingyu Zeng.
\newblock Zerosense: How vision matters in long context compression.
\newblock \emph{arXiv preprint arXiv:2603.11846}, 2026.

\bibitem[Hsieh et~al.(2024)Hsieh, Sun, Kriman, Acharya, Rekesh, Jia, Zhang, and Ginsburg]{hsieh2024ruler}
Cheng-Ping Hsieh, Simeng Sun, Samuel Kriman, Shantanu Acharya, Dima Rekesh, Fei Jia, Yang Zhang, and Boris Ginsburg.
\newblock Ruler: What's the real context size of your long-context language models?
\newblock \emph{arXiv preprint arXiv:2404.06654}, 2024.

\bibitem[Hu et~al.(2025)Hu, Xu, Zhang, Ye, Yan, Zhang, Jin, Huang, and Zhou]{hu2024docowl2}
Anwen Hu, Haiyang Xu, Liang Zhang, Jiabo Ye, Ming Yan, Ji~Zhang, Qin Jin, Fei Huang, and Jingren Zhou.
\newblock m{PLUG}-{D}oc{O}wl2: High-resolution compressing for {OCR}-free multi-page document understanding.
\newblock In Wanxiang Che, Joyce Nabende, Ekaterina Shutova, and Mohammad~Taher Pilehvar, editors, \emph{Proceedings of the 63rd Annual Meeting of the Association for Computational Linguistics (Volume 1: Long Papers)}, pages 5817--5834, Vienna, Austria, July 2025. Association for Computational Linguistics.
\newblock ISBN 979-8-89176-251-0.
\newblock \doi{10.18653/v1/2025.acl-long.291}.
\newblock URL \url{https://aclanthology.org/2025.acl-long.291/}.

\bibitem[Huang et~al.(2025)Huang, Radhakrishnan, Yu, and Kautz]{huang2025frag}
De-An Huang, Subhashree Radhakrishnan, Zhiding Yu, and Jan Kautz.
\newblock Frag: Frame selection augmented generation for long video and long document understanding.
\newblock \emph{arXiv preprint arXiv:2504.17447}, 2025.

\bibitem[Husain et~al.(2019)Husain, Wu, Gazit, Allamanis, and Brockschmidt]{husain2019codesearchnet}
Hamel Husain, Ho-Hsiang Wu, Tiferet Gazit, Miltiadis Allamanis, and Marc Brockschmidt.
\newblock {CodeSearchNet} challenge: Evaluating the state of semantic code search.
\newblock \emph{arXiv preprint arXiv:1909.09436}, 2019.
\newblock URL \url{https://arxiv.org/abs/1909.09436}.

\bibitem[Jain et~al.(2025)Jain, Wu, Zeng, Liu, Dai, Shao, Wu, and Wang]{jain2025simpledoc}
Chelsi Jain, Yiran Wu, Yifan Zeng, Jiale Liu, Shengyu Dai, Zhenwen Shao, Qingyun Wu, and Huazheng Wang.
\newblock Simpledoc: Multi-modal document understanding with dual-cue page retrieval and iterative refinement.
\newblock In \emph{Proceedings of the 2025 Conference on Empirical Methods in Natural Language Processing}, pages 28398--28415, 2025.

\bibitem[Jiang et~al.(2023)Jiang, Wu, Lin, Yang, and Qiu]{jiang2023llmlingua}
Huiqiang Jiang, Qianhui Wu, Chin-Yew Lin, Yuqing Yang, and Lili Qiu.
\newblock {LLML}ingua: Compressing prompts for accelerated inference of large language models.
\newblock In Houda Bouamor, Juan Pino, and Kalika Bali, editors, \emph{Proceedings of the 2023 Conference on Empirical Methods in Natural Language Processing}, pages 13358--13376, Singapore, December 2023. Association for Computational Linguistics.
\newblock \doi{10.18653/v1/2023.emnlp-main.825}.
\newblock URL \url{https://aclanthology.org/2023.emnlp-main.825/}.

\bibitem[Jiang et~al.(2025)Jiang, Xu, Lin, Xiao, Wang, Sun, and Han]{jiang-etal-2025-s3}
Pengcheng Jiang, Xueqiang Xu, Jiacheng Lin, Jinfeng Xiao, Zifeng Wang, Jimeng Sun, and Jiawei Han.
\newblock s3: You don{'}t need that much data to train a search agent via {RL}.
\newblock In Christos Christodoulopoulos, Tanmoy Chakraborty, Carolyn Rose, and Violet Peng, editors, \emph{Proceedings of the 2025 Conference on Empirical Methods in Natural Language Processing}, pages 21599--21617, Suzhou, China, November 2025. Association for Computational Linguistics.
\newblock ISBN 979-8-89176-332-6.
\newblock \doi{10.18653/v1/2025.emnlp-main.1095}.
\newblock URL \url{https://aclanthology.org/2025.emnlp-main.1095/}.

\bibitem[Jin et~al.(2025)Jin, Zeng, Yue, Yoon, Arik, Wang, Zamani, and Han]{jin2025search}
Bowen Jin, Hansi Zeng, Zhenrui Yue, Jinsung Yoon, Sercan Arik, Dong Wang, Hamed Zamani, and Jiawei Han.
\newblock Search-r1: Training llms to reason and leverage search engines with reinforcement learning.
\newblock \emph{arXiv preprint arXiv:2503.09516}, 2025.

\bibitem[Kwiatkowski et~al.(2019)Kwiatkowski, Palomaki, Redfield, Collins, Parikh, Alberti, Epstein, Polosukhin, Devlin, Lee, Toutanova, Jones, Kelcey, Chang, Dai, Uszkoreit, Le, and Petrov]{kwiatkowski2019nq}
Tom Kwiatkowski, Jennimaria Palomaki, Olivia Redfield, Michael Collins, Ankur Parikh, Chris Alberti, Danielle Epstein, Illia Polosukhin, Jacob Devlin, Kenton Lee, Kristina Toutanova, Llion Jones, Matthew Kelcey, Ming-Wei Chang, Andrew~M. Dai, Jakob Uszkoreit, Quoc Le, and Slav Petrov.
\newblock Natural questions: A benchmark for question answering research.
\newblock \emph{Transactions of the Association for Computational Linguistics}, 7:\penalty0 452--466, 2019.
\newblock \doi{10.1162/tacl_a_00276}.
\newblock URL \url{https://aclanthology.org/Q19-1026/}.

\bibitem[Kwon et~al.(2023)Kwon, Li, Zhuang, Sheng, Zheng, Yu, Gonzalez, Zhang, and Stoica]{kwon2023vllm}
Woosuk Kwon, Zhuohan Li, Siyuan Zhuang, Ying Sheng, Lianmin Zheng, Cody~Hao Yu, Joseph Gonzalez, Hao Zhang, and Ion Stoica.
\newblock Efficient memory management for large language model serving with pagedattention.
\newblock In \emph{Proceedings of the 29th symposium on operating systems principles}, pages 611--626, 2023.

\bibitem[Lewis et~al.(2020)Lewis, Perez, Piktus, Petroni, Karpukhin, Goyal, K{\"u}ttler, Lewis, Yih, Rockt{\"a}schel, et~al.]{lewis2020rag}
Patrick Lewis, Ethan Perez, Aleksandra Piktus, Fabio Petroni, Vladimir Karpukhin, Naman Goyal, Heinrich K{\"u}ttler, Mike Lewis, Wen-tau Yih, Tim Rockt{\"a}schel, et~al.
\newblock Retrieval-augmented generation for knowledge-intensive nlp tasks.
\newblock \emph{Advances in neural information processing systems}, 33:\penalty0 9459--9474, 2020.

\bibitem[Liu et~al.(2024)Liu, Tian, Daita, Wei, Ding, Wang, Yang, and Zhang]{liu2024repoqa}
Jiawei Liu, Jia~Le Tian, Vijay Daita, Yuxiang Wei, Yifeng Ding, Yuhan~Katherine Wang, Jun Yang, and Lingming Zhang.
\newblock Repoqa: Evaluating long context code understanding.
\newblock \emph{arXiv preprint arXiv:2406.06025}, 2024.

\bibitem[Ma et~al.(2024)Ma, Zang, Chen, Chen, Jiao, Li, Lu, Liu, Ma, Dong, et~al.]{ma2024mmlongbench}
Yubo Ma, Yuhang Zang, Liangyu Chen, Meiqi Chen, Yizhu Jiao, Xinze Li, Xinyuan Lu, Ziyu Liu, Yan Ma, Xiaoyi Dong, et~al.
\newblock Mmlongbench-doc: Benchmarking long-context document understanding with visualizations.
\newblock \emph{Advances in Neural Information Processing Systems}, 37:\penalty0 95963--96010, 2024.

\bibitem[Pan et~al.(2024)Pan, Wu, Jiang, Xia, Luo, Zhang, Lin, R{\"u}hle, Yang, Lin, Zhao, Qiu, and Zhang]{pan2024llmlingua2}
Zhuoshi Pan, Qianhui Wu, Huiqiang Jiang, Menglin Xia, Xufang Luo, Jue Zhang, Qingwei Lin, Victor R{\"u}hle, Yuqing Yang, Chin-Yew Lin, H.~Vicky Zhao, Lili Qiu, and Dongmei Zhang.
\newblock {LLML}ingua-2: Data distillation for efficient and faithful task-agnostic prompt compression.
\newblock In Lun-Wei Ku, Andre Martins, and Vivek Srikumar, editors, \emph{Findings of the Association for Computational Linguistics: ACL 2024}, pages 963--981, Bangkok, Thailand, August 2024. Association for Computational Linguistics.
\newblock \doi{10.18653/v1/2024.findings-acl.57}.
\newblock URL \url{https://aclanthology.org/2024.findings-acl.57/}.

\bibitem[{Qwen Team}(2026)]{qwen3.5}
{Qwen Team}.
\newblock {Qwen3.5} technical report.
\newblock \emph{https://qwen.ai/blog?id=qwen3.5}, 2026.

\bibitem[Rust et~al.(2022)Rust, Lotz, Bugliarello, Salesky, de~Lhoneux, and Elliott]{rust2023pixel}
Phillip Rust, Jonas~F Lotz, Emanuele Bugliarello, Elizabeth Salesky, Miryam de~Lhoneux, and Desmond Elliott.
\newblock Language modelling with pixels.
\newblock \emph{arXiv preprint arXiv:2207.06991}, 2022.

\bibitem[Sahu et~al.(2024)Sahu, Mandal, Bharadwaj, Kanade, Maniatis, and Shevade]{sahu2022codequeries}
Surya~Prakash Sahu, Madhurima Mandal, Shikhar Bharadwaj, Aditya Kanade, Petros Maniatis, and Shirish Shevade.
\newblock {CodeQueries}: A dataset of semantic queries over code.
\newblock In \emph{Proceedings of the 17th Innovations in Software Engineering Conference}, pages 1--11, 2024.

\bibitem[Sheng et~al.(2024)Sheng, Zhang, Ye, Wu, Zhang, Zhang, Peng, Lin, and Wu]{sheng2024verl}
Guangming Sheng, Chi Zhang, Zilingfeng Ye, Xibin Wu, Wang Zhang, Ru~Zhang, Yanghua Peng, Haibin Lin, and Chuan Wu.
\newblock Hybridflow: A flexible and efficient rlhf framework.
\newblock \emph{arXiv preprint arXiv: 2409.19256}, 2024.

\bibitem[Su et~al.(2025)Su, Xia, Guo, Liu, Ma, Qu, Liu, Li, Zeng, Yang, Li, Cheng, Ji, He, and Fung]{su2025twi_survey}
Zhaochen Su, Peng Xia, Hangyu Guo, Zhenhua Liu, Yan Ma, Xiaoye Qu, Jiaqi Liu, Yanshu Li, Kaide Zeng, Zhengyuan Yang, Linjie Li, Yu~Cheng, Heng Ji, Junxian He, and Yi~R. Fung.
\newblock Thinking with images for multimodal reasoning: Foundations, methods, and future frontiers.
\newblock \emph{arXiv preprint arXiv:2506.23918}, 2025.

\bibitem[Sun et~al.(2026)Sun, Yuan, Liu, Zhao, Zhang, Dredze, and Bai]{sun2026modality}
Kaiser Sun, Xiaochuang Yuan, Hongjun Liu, Chen Zhao, Cheng Zhang, Mark Dredze, and Fan Bai.
\newblock Reading, not thinking: Understanding and bridging the modality gap when text becomes pixels in multimodal llms.
\newblock \emph{arXiv preprint arXiv:2603.09095}, 2026.

\bibitem[Tay et~al.(2022)Tay, Dehghani, Bahri, and Metzler]{tay2022efficient}
Yi~Tay, Mostafa Dehghani, Dara Bahri, and Donald Metzler.
\newblock Efficient transformers: A survey.
\newblock \emph{ACM Computing Surveys}, 55\penalty0 (6):\penalty0 1--28, 2022.

\bibitem[Thapa et~al.(2024)Thapa, Chen, Covert, Chalamala, Athiwaratkun, Song, and Zou]{thapa2024dragonfly}
Rahul Thapa, Kezhen Chen, Ian Covert, Rahul Chalamala, Ben Athiwaratkun, Shuaiwen~Leon Song, and James Zou.
\newblock Dragonfly: Multi-resolution zoom-in encoding enhances vision-language models.
\newblock \emph{arXiv preprint arXiv:2406.00977}, 2024.

\bibitem[Tito et~al.(2023)Tito, Karatzas, and Valveny]{tito2022mpdocvqa}
Rub{\`e}n Tito, Dimosthenis Karatzas, and Ernest Valveny.
\newblock Hierarchical multimodal transformers for multipage docvqa.
\newblock \emph{Pattern Recognition}, 144:\penalty0 109834, 2023.

\bibitem[Trivedi et~al.(2022)Trivedi, Balasubramanian, Khot, and Sabharwal]{trivedi2022musique}
Harsh Trivedi, Niranjan Balasubramanian, Tushar Khot, and Ashish Sabharwal.
\newblock Musique: Multihop questions via single-hop question composition.
\newblock \emph{Transactions of the Association for Computational Linguistics}, 10:\penalty0 539--554, 2022.

\bibitem[Vaswani et~al.(2017)Vaswani, Shazeer, Parmar, Uszkoreit, Jones, Gomez, Kaiser, and Polosukhin]{vaswani2017attention}
Ashish Vaswani, Noam Shazeer, Niki Parmar, Jakob Uszkoreit, Llion Jones, Aidan~N Gomez, {\L}ukasz Kaiser, and Illia Polosukhin.
\newblock Attention is all you need.
\newblock \emph{Advances in neural information processing systems}, 30, 2017.

\bibitem[Wan et~al.(2025)Wan, Shen, Liao, Shi, Li, Yang, Zhang, Huang, Zhou, and Yan]{wan2025qwenlong}
Fanqi Wan, Weizhou Shen, Shengyi Liao, Yingcheng Shi, Chenliang Li, Ziyi Yang, Ji~Zhang, Fei Huang, Jingren Zhou, and Ming Yan.
\newblock Qwenlong-l1: Towards long-context large reasoning models with reinforcement learning.
\newblock \emph{arXiv preprint arXiv:2505.17667}, 2025.

\bibitem[Wang et~al.(2026)Wang, Jing, Liu, Guan, Tu, Wang, Huang, and Tao]{wang2026vtcr1}
Yibo Wang, Yongcheng Jing, Shunyu Liu, Hao Guan, Rong-cheng Tu, Chengyu Wang, Jun Huang, and Dacheng Tao.
\newblock Vtc-r1: Vision-text compression for efficient long-context reasoning.
\newblock \emph{arXiv preprint arXiv:2601.22069}, 2026.

\bibitem[Wei et~al.(2025)Wei, Sun, and Li]{wei2025deepseekocr}
Haoran Wei, Yaofeng Sun, and Yukun Li.
\newblock {DeepSeek-OCR}: Contexts optical compression.
\newblock \emph{arXiv preprint arXiv:2510.18234}, 2025.

\bibitem[Wu et~al.(2025)Wu, Yang, Jiang, Li, Yan, Yu, Zhang, Zhai, and Nahrstedt]{wu2026vtoolr1}
Mingyuan Wu, Jingcheng Yang, Jize Jiang, Meitang Li, Kaizhuo Yan, Hanchao Yu, Minjia Zhang, Chengxiang Zhai, and Klara Nahrstedt.
\newblock Vtool-r1: Vlms learn to think with images via reinforcement learning on multimodal tool use, 2025.
\newblock URL \url{https://arxiv.org/abs/2505.19255}.

\bibitem[Wu et~al.(2024)Wu, Wang, Xiao, Peng, and Fu]{wu2024retrieval}
Wenhao Wu, Yizhong Wang, Guangxuan Xiao, Hao Peng, and Yao Fu.
\newblock Retrieval head mechanistically explains long-context factuality.
\newblock \emph{arXiv preprint arXiv:2404.15574}, 2024.

\bibitem[Xiao et~al.(2023)Xiao, Tian, Chen, Han, and Lewis]{xiao2024efficient}
Guangxuan Xiao, Yuandong Tian, Beidi Chen, Song Han, and Mike Lewis.
\newblock Efficient streaming language models with attention sinks.
\newblock \emph{arXiv}, 2023.

\bibitem[Xie et~al.(2025)Xie, Wang, Rosu, Deng, Sun, Lin, and Dhingra]{xie2025when}
Roy Xie, Junlin Wang, Paul Rosu, Chunyuan Deng, Bolun Sun, Zihao Lin, and Bhuwan Dhingra.
\newblock Knowing when to stop: Efficient context processing via latent sufficiency signals.
\newblock \emph{Advances in Neural Information Processing Systems}, 2025.

\bibitem[Xing et~al.(2025)Xing, Wang, Yan, Shu, and Tang]{xing2025vist}
Ling Xing, Alex~Jinpeng Wang, Rui Yan, Xiangbo Shu, and Jinhui Tang.
\newblock Vision-centric token compression in large language model.
\newblock \emph{arXiv preprint arXiv:2502.00791}, 2025.

\bibitem[Yang et~al.(2018)Yang, Qi, Zhang, Bengio, Cohen, Salakhutdinov, and Manning]{yang2018hotpotqa}
Zhilin Yang, Peng Qi, Saizheng Zhang, Yoshua Bengio, William Cohen, Ruslan Salakhutdinov, and Christopher~D. Manning.
\newblock {H}otpot{QA}: A dataset for diverse, explainable multi-hop question answering.
\newblock In Ellen Riloff, David Chiang, Julia Hockenmaier, and Jun{'}ichi Tsujii, editors, \emph{Proceedings of the 2018 Conference on Empirical Methods in Natural Language Processing}, pages 2369--2380, Brussels, Belgium, October-November 2018. Association for Computational Linguistics.
\newblock \doi{10.18653/v1/D18-1259}.
\newblock URL \url{https://aclanthology.org/D18-1259/}.

\bibitem[Yen et~al.(2025)Yen, Gao, Hou, Ding, Fleischer, Izsak, Wasserblat, and Chen]{yen2024helmet}
Howard Yen, Tianyu Gao, Minmin Hou, Ke~Ding, Daniel Fleischer, Peter Izsak, Moshe Wasserblat, and Danqi Chen.
\newblock Helmet: How to evaluate long-context language models effectively and thoroughly.
\newblock In \emph{International Conference on Learning Representations (ICLR)}, 2025.

\bibitem[Yu et~al.(2025)Yu, Zhang, Zhu, Yuan, Zuo, Yue, Dai, Fan, Liu, Liu, et~al.]{yu2025dapo}
Qiying Yu, Zheng Zhang, Ruofei Zhu, Yufeng Yuan, Xiaochen Zuo, Yu~Yue, Weinan Dai, Tiantian Fan, Gaohong Liu, Lingjun Liu, et~al.
\newblock Dapo: An open-source llm reinforcement learning system at scale.
\newblock \emph{arXiv preprint arXiv:2503.14476}, 2025.

\bibitem[Zhang et~al.(2025)Zhang, Li, Long, Zhang, Lin, Yang, Xie, Yang, Liu, Lin, Huang, and Zhou]{zhang2025qwen3embedding}
Yanzhao Zhang, Mingxin Li, Dingkun Long, Xin Zhang, Huan Lin, Baosong Yang, Pengjun Xie, An~Yang, Dayiheng Liu, Junyang Lin, Fei Huang, and Jingren Zhou.
\newblock {Qwen3 Embedding}: Advancing text embedding and reranking through foundation models.
\newblock \emph{arXiv preprint arXiv:2506.05176}, 2025.
\newblock URL \url{https://arxiv.org/abs/2506.05176}.

\bibitem[Zheng et~al.(2024)Zheng, Zhang, Zhang, Ye, Luo, Feng, and Ma]{zheng2024llamafactory}
Yaowei Zheng, Richong Zhang, Junhao Zhang, Yanhan Ye, Zheyan Luo, Zhangchi Feng, and Yongqiang Ma.
\newblock Llamafactory: Unified efficient fine-tuning of 100+ language models.
\newblock In \emph{Proceedings of the 62nd Annual Meeting of the Association for Computational Linguistics (Volume 3: System Demonstrations)}, Bangkok, Thailand, 2024. Association for Computational Linguistics.
\newblock URL \url{http://arxiv.org/abs/2403.13372}.

\bibitem[Zheng et~al.(2025)Zheng, Yang, Hong, Zhao, Xu, Yang, Shen, and Yu]{zheng2026deepeyes}
Ziwei Zheng, Michael Yang, Jack Hong, Chenxiao Zhao, Guohai Xu, Le~Yang, Chao Shen, and Xing Yu.
\newblock Deepeyes: Incentivizing" thinking with images" via reinforcement learning.
\newblock \emph{arXiv preprint arXiv:2505.14362}, 2025.

\end{thebibliography}
